\documentclass[letterpaper]{article} % DO NOT CHANGE THIS
\usepackage{aaai25}  % DO NOT CHANGE THIS
\usepackage{times}  % DO NOT CHANGE THIS
\usepackage{helvet}  % DO NOT CHANGE THIS
\usepackage{courier}  % DO NOT CHANGE THIS
\usepackage[hyphens]{url}  % DO NOT CHANGE THIS
\usepackage{graphicx} % DO NOT CHANGE THIS
\usepackage{natbib}  % DO NOT CHANGE THIS AND DO NOT ADD ANY OPTIONS TO IT
\usepackage{caption} % DO NOT CHANGE THIS AND DO NOT ADD ANY OPTIONS TO IT
\usepackage{algorithm}
\usepackage{newfloat}
\usepackage{listings}
\DeclareCaptionStyle{ruled}{labelfont=normalfont,labelsep=colon,strut=off} % DO NOT CHANGE THIS
\floatstyle{ruled}
\newfloat{listing}{tb}{lst}{}
\floatname{listing}{Listing}
\nocopyright
\usepackage{amsmath}
\usepackage{amssymb}
\usepackage{mathtools}
\usepackage{amsthm}
\usepackage{graphicx}
\usepackage{appendix}
\usepackage{paralist}
\usepackage{booktabs} 
\usepackage{algorithm}
\usepackage{algpseudocode}

\usepackage{makecell}
\usepackage{multirow}
\usepackage{enumitem}
\usepackage{subcaption}
\usepackage{appendix}
\usepackage{paralist}
\usepackage{bbding}
\usepackage{pifont}
\usepackage{xcolor}

\newtheorem{assumption} {Assumption}

\newtheorem{theorem}{Theorem}

\newtheorem{lemma}{Lemma}

\newtheorem{definition}{Definition}

\def \bx        {\mathbf{x}}

\def \tf        {\Tilde{f}}
\def \hf        {\widehat{f}}

\def \hy        {\hat{\mathbf{y}}}

\def \y         {\mathbf{y}}
\def \x         {\mathbf{x}}

\def \u         {\mathbf{u}}
\def \v         {\mathbf{v}}

\def \ts        {s}

\def \ty        {\Tilde{\mathbf{y}}}

\def \B         {\mathcal{B}}

\def \O         {\mathcal{O}}
\def \K         {\mathcal{K}}

\def \E         {\mathbb{E}}

\def \hf        {\hat{f}}

\def \tG        {\Tilde{G}}
\def \tQ        {\beta Q}
\def \tnabla    {\Tilde{\nabla}}
\def \hnabla    {\hat{\nabla}}

\DeclareMathOperator*{\argmin}{argmin}

\def \ta        {\mathtt{term~(a)}}
\def \tb        {\mathtt{term~(b)}}
\def \tc        {\mathtt{term~(c)}}
\def \td        {\mathtt{term~(d)}}
\def \te        {\mathtt{term~(e)}}
\def \termf     {\mathtt{term~(f)}}
\def \termg     {\mathtt{term~(g)}}
\def \termh     {\mathtt{term~(h)}}
\def \termi     {\mathtt{term~(i)}}

\def \sumT      {\sum_{t=1}^T}

\def \max       {\mathrm{max}}

\def \Reg       {\mathrm{Regret}}

\title{Revisiting Projection-Free Online Learning  with Time-Varying Constraints}
\author{
    Yibo Wang\textsuperscript{\rm 1,2}, \
    Yuanyu Wan\textsuperscript{\rm 3,1,}\footnotemark[1], \
    Lijun Zhang\textsuperscript{\rm 1,2,}\thanks{  Corresponding author}
}
\affiliations{
    \textsuperscript{\rm 1} National Key Laboratory for Novel Software Technology, Nanjing University, Nanjing 210023, China\\
    \textsuperscript{\rm 2} School of Artificial Intelligence, Nanjing University, Nanjing 210023, China\\
    \textsuperscript{\rm 3} School of Software Technology, Zhejiang University, Ningbo 315048, China\\
    \{wangyb, zhanglj\}@lamda.nju.edu.cn, wanyy@zju.edu.cn
}

\usepackage{bibentry}
\begin{document}

\maketitle

\begin{abstract}
We investigate constrained online convex optimization, in which decisions must belong to a fixed and typically complicated domain, and are required to approximately satisfy additional time-varying constraints over the long term. In this setting, the commonly used projection operations are often computationally expensive or even intractable. To avoid the time-consuming operation, several projection-free methods have been proposed with an $\mathcal{O}(T^{3/4} \sqrt{\log T})$ regret bound and an $\mathcal{O}(T^{7/8})$ cumulative constraint violation (CCV) bound for general convex losses. In this paper, we improve this result and further establish \textit{novel} regret and CCV bounds when loss functions are strongly convex. The primary idea is to first construct a composite surrogate loss, involving the original loss and constraint functions, by utilizing the Lyapunov-based technique. Then, we propose a parameter-free variant of the classical projection-free method, namely online Frank-Wolfe (OFW), and run this new extension over the online-generated surrogate loss. Theoretically, for general convex losses, we achieve an $\mathcal{O}(T^{3/4})$ regret bound and an $\mathcal{O}(T^{3/4} \log T)$ CCV bound, both of which are order-wise tighter than existing results. For strongly convex losses, we establish new guarantees of an $\mathcal{O}(T^{2/3})$ regret bound and an $\mathcal{O}(T^{5/6})$ CCV bound. Moreover, we also extend our methods to a more challenging setting with bandit feedback, obtaining similar theoretical findings. Empirically, experiments on real-world datasets have demonstrated the effectiveness of our methods.
\end{abstract}


\section{Introduction}
Online convex optimization (OCO) has become a popular paradigm for modeling online decision-making problems \cite{Others:2012:Shalev-Shwartz,Others:2016:Hazan,ArXiv:2019:Orabona}, such as  online portfolio optimization \cite{ICML:2006:Agarwal} and online advertisement system \cite{Others:2013:McMahan}. Formally, OCO can be viewed as a structured iterative game between a learner and an adversary.  Specifically, at each round $t$, the learner first chooses a decision $\x_t$ from a convex and fixed domain $\K \subseteq \mathbb{R}^d$. Then, the adversary reveals a convex loss function $f_t(\cdot): \K  \mapsto \mathbb{R}$, and the learner suffers the cost $f_t(\x_t)$. The goal of the learner is to minimize the regret:
\begin{equation}
    \label{eq:regret}
    \Reg_T = \sum\nolimits_{t=1}^T f_t(\x_t) - \min_{\x \in \K} \sum\nolimits_{t=1}^T f_t(\x),  
\end{equation}
defined as the difference between the cumulative loss of the learner and that of the best fixed decision. 

In the literature,  there have been abundant theoretical appeals for OCO, such as the $\O(\sqrt{T})$ regret bound for general convex losses \cite{ICML:2003:Zinkevich} and the $\O(\log T)$ regret bound for strongly convex losses \cite{Others:2007:Hazan}. In practice, besides the \textit{hard and fixed} domain $\K$,  the decisions made by the learner are typically governed by a series of \textit{soft and time-varying} constraints, which may be violated in several rounds but should be satisfied on average over the long term. For example, in wireless communication systems, operators manage varying transmission power consumption to ensure the reception of messages \cite{JMLR:2009:Mannor}; in online advertisement systems, advertisers   employ  dynamic  budgets to maximize the click-through-rates for their advertisements \cite{ICML:2019:Liakopoulos}. These practical applications thus motivate the development of constrained online convex optimization (COCO) \cite{JMLR:2012:Mahdavi,ArXiv:2017:Neely}. 

In the framework of COCO, the time-varying constraints are typically captured by the inequality $g_t(\x) \leq 0$ where $g_t(\cdot): \K  \mapsto \mathbb{R}$ is a convex function revealed by the adversary at the end of each round $t$. Consequently, in addition to minimizing   \eqref{eq:regret}, the learner also aims to ensure the cumulative constraint violation (CCV):
\begin{equation}
    \label{eq:ccv}
    Q_T = \sum\nolimits_{t=1}^T g_t^+(\x_t)
\end{equation}
to be sublinear with respect to the time horizon $T$, where $g_t^+(\x) \triangleq \max\{0, g_t(\x)\}$. To optimize~\eqref{eq:regret}~and~\eqref{eq:ccv} concurrently,  various efforts have been made recently \cite{TAC:2019:Cao,JMLR:2020:Yu,ICML:2021:Yi,NeurIPS:2022:Guo,TAC:2023:Yi,NeurIPS:2024:Sinha}, and established plentiful guarantees, including the regret and CCV bounds of $\O(\sqrt{T})$  for general convex losses \cite{NeurIPS:2017:Yu}.

The key operation in these COCO methods is the projection that pulls an infeasible decision back into the hard constraint $\K$. In many practical scenarios, the domain $\K$ is typically high-dimensional and complex, rendering projections onto  $\K$ computationally expensive or even intractable, which significantly limits the applicability of these methods. To address this issue, several studies \cite{ArXiv:2023:Lee,ICML:2024:Garber} propose projection-free methods for COCO, which replace the time-consuming projection with the more efficient linear optimization operation. One prominent example is online semidefinite optimization, where the hard constraint is a positive semidefinite cone with the bounded trace. In this case, the linear optimization has been proven at least an order of magnitude faster than   projection  \cite{ICML:2012:Hazan}. Unfortunately, existing projection-free methods  can only  guarantee the      $\O(T^{3/4} \sqrt{\log T})$ regret bound  and the    $\O(T^{7/8})$ CCV bound for general convex losses \cite{ICML:2024:Garber}.

In this paper, we improve the above bounds and introduce \textit{new} theoretical guarantees for strongly convex losses. The key idea is to first construct a composite surrogate loss that consists of the original loss $f_t(\cdot)$ and the time-varying constraint $g_t(\cdot)$, based on a carefully designed Lyapunov function. Rigorous analysis reveals that both   \eqref{eq:regret} and \eqref{eq:ccv} are simultaneously controlled by the regret in terms of the surrogate losses, so that we can directly apply classical projection-free methods, e.g.,~online Frank-Wolfe (OFW) \cite{ICML:2012:Hazan}, over the surrogate losses to minimize the two metrics. Notably, since the surrogate loss  is generated in an online manner, essential prior knowledge for OFW (e.g.,~the gradient norm bound) is unavailable beforehand. Therefore, we need to employ the methods that are agnostic to the prior parameters about the surrogate loss. To this end, we propose the \textit{first} parameter-free variant of OFW for general convex losses based on the doubling trick technique \cite{Others:1997:Bianchi}. By running the parameter-free variant over the composite surrogate losses, we establish an $\O(T^{3/4})$ regret bound and an $\O(T^{3/4} \log T)$ CCV bound for the general convex loss.  Both of our results are better than  the state-of-the-art bounds achieved by \citet{ICML:2024:Garber}. Additionally, we further investigate the strongly convex loss and achieve  an $\O(T^{2/3})$  regret bound and an $\O(T^{5/6})$ CCV bound, by constructing the surrogate loss based on a different Lyapunov function and running the strongly convex variant of  OFW \cite{AAAI:2021:Wan}.

Furthermore, to handle the more challenging bandit setting, we combine our proposed methods with the classical one-point estimator \cite{SODA:2005:Flaxman}, which can approximate the gradient with only the loss value. Theoretically, for general convex losses, we  establish the  $\O(T^{3/4})$ regret bound and the  $\O(T^{3/4} \log T)$ CCV bound. For strongly convex losses,  we achieve the $\O(T^{2/3} \log T)$ regret bound and the $\O(T^{5/6} \log T)$ CCV bound.

\paragraph{Contributions.} We summarize our contributions below.
\begin{itemize}
    \item For general convex losses, we deliver an $\O(T^{3/4})$ regret bound and an $\O(T^{3/4} \log T)$ CCV bound, both of which improve the previous results of \citet{ICML:2024:Garber}. During the analysis, we propose the \textit{first} parameter-free variant of OFW, which may be an independent of interest;

    \item For  strongly convex losses, we establish the \textit{novel} results of an  $\O(T^{2/3})$  regret bound and an $\O(T^{5/6})$ CCV bound for projection-free COCO;

    \item We extend our methods to the bandit setting and achieve similar bounds as those in the full-information setting;

    \item We verify our theoretical findings by conducting experiments on real-world datasets. The empirical results have   demonstrated the effectiveness  of our methods.

\end{itemize}

\section{Related Work}

\begin{table*}[t]
    \centering
        \begin{tabular}{c|ccc|cc}
            \toprule
            \textbf{Methods} & \textbf{Losses} &   \textbf{Constraints} & \textbf{Feedback} & \textbf{Regret} & \textbf{CCV} \\ \midrule
              \multicolumn{1}{c|}{\multirow{2}{*}{\citet{ArXiv:2023:Lee}}}  & cvx &  sto  & full-info &  $\O(T^{5/6})$  & $\O(T^{5/6})$ \\  
               &   cvx  & adv  & full-info & $\O(T^{5/6 + \alpha})$ & $\O(T^{11/12 - \alpha/2})$ \\  
             \citet{ICML:2024:Garber}  & cvx  & adv  & full-info & $\O(T^{3/4} \sqrt{\log T})$    & $\O(T^{7/8})$ \\   
             \textbf{Theorem~\ref{thm:cvx}} (this work)  & cvx   & adv  &  full-info &  $\O(T^{3/4})$ & $\O(T^{3/4} \log T)$ \\ 
             \textbf{Theorem~\ref{thm:str:f}} (this work)  & str-cvx &   adv  &  full-info &  $\O(T^{2/3})$ & $\O(T^{5/6})$ \\ 
              \midrule
             \citet{ICML:2024:Garber}  & cvx  & adv  & bandits & $\O(T^{3/4} \sqrt{\log T})$    & $\O(T^{7/8} \log T  )$ \\  
             \textbf{Theorem~\ref{thm:bdt:cvx}} (this work)  & cvx   & adv  &  bandits &  $\O(T^{3/4})$ & $\O(T^{3/4} \log T)$ \\ 
             \textbf{Theorem~\ref{thm:bdt:str:f}} (this work)  & str-cvx  & adv  &  bandits &  $\O(T^{2/3} \log T)$ & $\O(T^{5/6} \log T)$ \\ 
             \bottomrule
        \end{tabular}
     % }
    \caption{Comparisons of our results with existing projection-free methods for COCO. Abbreviations: convex $\rightarrow$ cvx, strongly convex $\rightarrow$ str-cvx, stochastic $\rightarrow$ sto, adversarial $\rightarrow$ adv, full-information $\rightarrow$ full-info.}
    \label{tab:related:work}
\end{table*}

In this section, we briefly overview the recent progress on projection-free and constrained online convex optimization.

\subsection{Projection-Free Online Convex Optimization}
The pioneering work of \citet{ICML:2012:Hazan} introduces the first projection-free online method, namely online Frank-Wolfe (OFW), which is an online extension of  the classical   Frank-Wolfe algorithm \cite{Others:1956:Frank}. The basic idea is to replace the time-consuming projection with the following linear optimization steps:
\begin{equation*}
        \label{eq:OFW}
         \v_t = \underset {\x \in \mathcal{K}} { \operatorname {argmin} }  \left < \nabla F_t(\x_t), \bx\right >,~
        \x_{t+1}  = \x_t + \sigma_t(\v_t - \x_t) 
\end{equation*}
where $\sigma_t > 0$ denotes the step size, and  $F_t(\x)$ is defined as
$$
    F_t(\x) = \eta \sum\nolimits_{\tau=1}^{t-1}\nabla f_\tau (\x_\tau)^\top \x + \| \x - \x_1\|_2^2
$$
parameterized by $\eta > 0$. With the prior knowledge about $f_t(\cdot)$ (e.g.,~the gradient norm bound) and appropriate configurations on $\eta$ and $\sigma_t$, OFW ensures an $\O(T^{3/4})$ regret bound for general convex losses. 

Based on OFW, plenty of investigations deliver tighter regret bounds by utilizing additional properties on $f_t(\cdot)$, such as the smoothness \cite{COLT:2020:Hazan}, the strong convexity \cite{AAAI:2021:Wan,AISTATS:2021:Kretzu}, and the exponential concavity \cite{COLT:2023:Garber,ArXiv:2024:Mhammedi}. Moreover, several efforts improve the regret bounds by  leveraging  special structures of    $\mathcal{K}$ \cite{Others:2016:Garber,AISTATS:2019:Levy,ArXiv:2020:Molinaro,AAAI:2021:Wan,COLT:2022:Mhammedi,NeurIPS:2023:Gatmiry}. Additionally, there exist other studies exploring  more practical scenarios, e.g.,~the bandit feedback \cite{AISTATS:2019:Chen,AISTATS:2020:Garber,PAKDD:2024:Zhang}, the delayed feedback \cite{ArXiv:2022:Wan}, the distributed setting \cite{ICML:2017:Zhang,ICML:2020:Wan,JMLR:2022:Wan,AAAI:2023:Wang,ArXiv:2024:Wan} and non-stationary environments \citep{Others:2021:Kalhan,AAAI:2021:Wan:B,COLT:2022:Garber,ALT:2023:Lu,COLT:2023:Wan,AAAI:2024:Wang}.

\subsection{Constrained Online Convex Optimization}
In the literature, there are two lines of research for COCO. One is the time-invariant setting, where the soft constraints are assumed to be fixed, i.e.,~$g_t(\cdot) = g(\cdot)$, and known to the learner at the beginning round. In this setting, for general convex losses, \citet{JMLR:2012:Mahdavi} originally develop  an $\O(\sqrt{T})$ regret bound and an $\O(T^{3/4})$ CCV bound. Then, subsequent studies generalize the results, and obtain tighter bounds for both regret and CCV under additional conditions \cite{ICML:2016:Jenatton,NeurIPS:2018:Yuan,JMLR:2020:Yu,ICML:2021:Yi}.

The other is the time-variant setting, where the soft constraints change over time and are only revealed after the learner submits the decision. Under the stochastic  time-varying constraints and the Slater's condition,  \citet{NeurIPS:2017:Yu} deliver an $\O(\sqrt{T})$ regret bound and an  $\O(\sqrt{T})$ CCV bound. Subsequently, extensive  studies focus on the more general adversarial time-varying constraints and attempt to remove the Slater's condition \cite{ArXiv:2017:Neely,ICML:2017:Sun,ICML:2019:Liakopoulos,TAC:2019:Cao,NeurIPS:2022:Guo,TAC:2023:Yi,NeurIPS:2024:Sinha}. One of the key techniques in these work is to analyze a  refined bound based on the Lyapunov drift of a virtual queue, which partially inspires our methods. To the best of our knowledge, the state-of-the-art results in this setting are delivered by \citet{NeurIPS:2024:Sinha}, who  establish the $\O(\sqrt{T})$ regret bound and the  $\O(\sqrt{T} \log T)$ CCV bound for general convex losses, and the $\O(\log T)$ regret bound and the  $\O(\sqrt{T \log T} )$ CCV bound for strongly convex losses.

As mentioned before, the above methods still rely on the inefficient projection for decision updates, which thereby motivates the development of projection-free COCO. \citet{ArXiv:2023:Lee}  first obtain an $\O(T^{5/6 + \alpha})$ regret bound and an $\O(T^{11/12-\alpha/2})$ CCV bound with the parameter $\alpha \in (0,1)$ for general convex losses and the full-information feedback. Later, \citet{ICML:2024:Garber} propose to apply a recent projection-free method, named LOO-BOGD \cite{COLT:2022:Garber}, under the   drift-plus-penalty framework \cite{Others:2010:Neely} that is extensively used in previous COCO methods \cite{NeurIPS:2017:Yu,NeurIPS:2022:Guo}, and thus deliver an $\O(T^{3/4}\sqrt{\log T})$ regret bound and an $\O(T^{7/8})$ CCV bound. When only the bandit feedback (i.e.,~the function value) is accessible, they obtain the same regret bound and a slightly worse   $\O(T^{7/8}\log T)$ CCV bound. More details can be found in Table~\ref{tab:related:work}.

\section{Preliminaries}

In this section, we recall   the basic assumptions and  definitions that are commonly used in prior  studies   \citep{JMLR:2012:Mahdavi,ICML:2012:Hazan,Others:2014:Agrawal}.

\begin{assumption}
    \label{assump:K-bound}
    The convex decision set $\mathcal{K}$ contains the ball of radius $r$ centered at the origin $\mathbf{0}$, and is contained in an   ball with the diameter $D = 2R$, i.e.,~$r \B \subseteq \mathcal{X} \subseteq R \B$ where $\B =\left\{\mathbf{x} \in \mathbb{R}^{d} \mid\|\mathbf{x}\|_{2} \leq 1\right\}$.
\end{assumption}

\begin{assumption}
    \label{assump:Lipschitz}
    At each round $t$, the loss function $f_{t}(\cdot)$ and the constraint function $g_t(\cdot)$ are $G$-Lipschitz over $\K$, i.e.,~$\forall \x, \y \in \K, |f_{t}(\x) - f_{t}(\y)| \leq   G \| \x - \y\|_2$ and $|g_{t}(\x) - g_{t}(\y)| \leq   G \| \x - \y\|_2$.
\end{assumption}

\begin{assumption}
    \label{assump:f-bound}
    At each round $t$, the loss function value $f_{t}(\bx)$ is bounded over $\K$, i.e.,~$\forall \bx \in \mathcal{K}, ~ |f_t(\bx)| \leq M.$
\end{assumption}

\begin{definition}
    \label{def:Lyapunov}
    Let $\Phi(x): \mathbb{R}^+ \mapsto \mathbb{R}$ be a convex function. It is called Lyapunov if $\Phi(x)$ satisfies (i) $\Phi(0)=0$; (ii) $\Phi(x) \geq 0, \forall x \in \mathbb{R}^+$; (iii) $\Phi(x)$ is non-decreasing. 
\end{definition}

\begin{definition} 
    \label{def:strong} 
     Let $f(\x): \mathcal{K} \mapsto \mathbb{R}$ be a function over $\mathcal{K}$. It is called $\alpha_f$-strongly convex if for all $\x, \y \in \mathcal{K}$
    \begin{equation} 
    f(\y) \geq f(\x) +  \langle \nabla f(\x), \y -\x  \rangle +  \frac{\alpha_f}{2}  \|\y -\x \|_2^2. \nonumber
    \end{equation}
\end{definition}
\noindent
In analysis, we will make use of the following property of strongly convex functions \cite{ICML:2015:Garber}.
\begin{lemma}       \label{lem:property:str}
    Let $f(\x)$ be an $\alpha_f$-strongly convex function over $\K$ and $\x^* = \argmin_{\x \in \K}f(\x)$. Then, for any $\x \in \K$
    \begin{equation}    \label{eq:property:str}
        \frac{\alpha_f}{2} \left\|\mathbf{x}-\mathbf{x}^{*}\right\|_{2}^{2} \leq f(\mathbf{x})-f\left(\mathbf{x}^{*}\right).
    \end{equation}
\end{lemma}

\section{Main Results}
In this section, we initially present our methods as well as their theoretical guarantees for the full-information setting. Then, we extend our investigations to the bandit setting. Due to the limitation of space, all proofs are deferred in the supplementary material.

\subsection{Algorithms for Full-Information Setting}

\begin{algorithm} [t]
    \caption{Online Frank-Wolfe with Time-Varying Constraints (OFW-TVC)}
    \label{alg1} 
    \textbf{Input:} Hyper-parameters $\beta$, $\gamma$, and the function $\Phi(\cdot)$ 
    \begin{algorithmic}[1]
        \State  Choose any $\x_1 \in \K$, and set $\Tilde{G}_1 = s_1 = k = 1$     \label{alg1:1}
        \For{$t = 1$ to $T$}
            \State  Play $\x_t$, and suffer $f_t(\x_t)$ and $g_t(\x_t)$  \label{alg1:3}
            \State Construct $Q_t$ and  $\tf_t(\x)$ according to \eqref{eq:Q} and \eqref{eq:tf}    \label{alg1:4}
            \While{$\tG_k < \beta G (\gamma +  \Phi'(\tQ_t))$} \label{alg1:5}
                    \State Set $\tG_{k+1} = 2\tG_k$, $\ts_{k+1} = t$, $k = k+1$  \label{alg1:6}
            \EndWhile    \label{alg1:7}
            \State Set $\eta_k$ and $F_{\ts_k:t}(\x)$ according to \eqref{eq:eta} and \eqref{eq:F}    \label{alg:line:eta} \label{alg1:8}
            
            \State Compute $\v_t$ and $\sigma_{\ts_k,t}$ according to \eqref{eq:v} and \eqref{eq:sigma}  \label{alg1:9}

            \State Update $\x_{t+1}$ according to  \eqref{eq:update:x}     \label{alg1:10}
        \EndFor
    \end{algorithmic}
\end{algorithm}

Overall, we first construct a composite surrogate loss function based on the loss $f_t(\x)$, the constraint $g_t(\x)$ and a specially designed  Lyapunov function that depends on the type of $f_t(\x)$. Then, we employ parameter-free variants of OFW to optimize the surrogate loss. 

Specifically,  let $ Q_t$ be the cumulative constraint violation at the round $t$, and $\Phi(\cdot)$ be a convex Lyapunov function. According to  \eqref{eq:ccv},  $Q_t$ can be formalized recursively as
\begin{equation}
    \label{eq:Q}
    Q_t =  Q_{t-1} +  g_t^+(\x_t),~\forall t \geq 1
\end{equation}
and $Q_0 = 0$.  By utilizing the convexity of $\Phi(\cdot)$, the Lyapunov drift of $Q_t$  at the round $t$, i.e.,~$\Phi(\beta Q_t) - \Phi(\beta Q_{t-1})$, is upper bounded by 
\begin{equation}
    \label{eq:phi}
    \begin{split}
        \Phi(\beta Q_t) - \Phi(\beta Q_{t-1}) \leq & \Phi'(\beta Q_t) \beta [ Q_t -  Q_{t-1}] \\
        \overset{\eqref{eq:Q}}{=} &   \Phi'(\beta Q_t) \beta  g_t^+(\x_t)
    \end{split}
\end{equation}
where $\beta > 0$ denotes a hyper-parameter. To simultaneously minimize $f_t(\x)$ and $g_t(\x)$, we follow the drift-plus-penalty framework \cite{Others:2010:Neely}, and construct the surrogate loss function $\tf_t(\x)$ by combining  the loss $f_t(\x)$ and  the   upper bound of the Lyapunov drift in \eqref{eq:phi}:
\begin{equation}    \label{eq:tf}
    \begin{split}
        \tf_t(\x)  = & \gamma \beta f_t(\x) +  \Phi'(\tQ_t) \beta g_t^+(\x)
    \end{split}
\end{equation}
where $\gamma > 0$ denotes a hyper-parameter. In fact, it can be verified that the regret in terms of $\tf_t(\x)$, denoted by $\Reg_{T}'$, concurrently captures \eqref{eq:regret} and \eqref{eq:ccv} \cite{NeurIPS:2024:Sinha}: 
\begin{equation}    \label{eq:regret_trans}
    \begin{split}
        \Reg_{T}'  
          \overset{\eqref{eq:phi}, \eqref{eq:tf}}{\geq} & \gamma \beta \Reg_{T} + \Phi(\tQ_T).
    \end{split}
\end{equation}
With an appropriate configuration on $\Phi(\cdot)$, \eqref{eq:regret} and \eqref{eq:ccv} can be  decoupled from \eqref{eq:regret_trans}, delivering corresponding theoretical guarantees. It should be noticed that the specific choice of  $\Phi(\cdot)$ is quite involved, since (i) it is employed to construct the surrogate loss in \eqref{eq:tf}, necessitating  a simple form that does not incur expensive computational costs; (ii) it appears in \eqref{eq:regret_trans} and is required to adeptly  balance the regret and CCV.

In the following, we investigate the general convex losses and the strongly convex losses.
\paragraph{General Convex Losses.} Given the favorable property of minimizing $\tf_t(\x)$ shown in \eqref{eq:regret_trans}, one may attempt to apply the classical OFW method over  $\tf_t(\x)$ for the simultaneous minimization on \eqref{eq:regret} and \eqref{eq:ccv}. However, such a straightforward application is not suitable, since  $\tf_t(\x)$ is generated in an online manner, and thus the prior knowledge  required by OFW  is unavailable beforehand. For example, the $\ell_2$-norm of the subgradient $\nabla \tf_t(\x)$ is bounded by:
\begin{equation}    \label{eq:tf:G}
    \begin{split}    
        \| \nabla \tf_t(\x) \|_2 \leq & \gamma \beta \| \nabla f_t(\x) \|_2  +  \Phi'(\tQ_t) \beta \|\nabla g_t(\x)\|_2 \\
        \leq &    \beta G (\gamma +  \Phi'(\tQ_T)) \triangleq \Tilde{G},
    \end{split}
\end{equation}
in which the last step follows the fact that $\Phi(\cdot)$ is convex and hence its derivative $\Phi'(\cdot)$ is non-decreasing. From \eqref{eq:tf:G}, it can be observed that $\Tilde{G}$ is unknown due to the uncertainty of $Q_T$ at the round $t$. For this reason, we propose the first parameter-free variant of OFW, which is agnostic to $\Tilde{G}$, and thereby can be employed to minimize $\tf_t(\x)$. The basic idea is to utilize an estimation of $\Tilde{G}$ for decision updating.  If the estimation is too low, we repeatedly  double the current guess and employ the first valid value for updates. We summarize our method in Algorithm~\ref{alg1}.

Specifically, at the Step~\ref{alg1:1}, we choose any point $\x_1 \in \K$ as the decision for the first round and make the estimation $\Tilde{G}_1=1$. Then, at each round $t$, we submit the decision $\x_t$, suffer the cost   $f_t(\x_t)$ and the constraint  $g_t(\x_t)$ (Step~\ref{alg1:3}). At the Step~\ref{alg1:4}, we construct $Q_t$ and the surrogate loss function $\tf_t(\x)$ according to \eqref{eq:Q} and \eqref{eq:tf}, respectively. Next, we verify the feasibility of the estimation $\Tilde{G}_k$. If it is  lower than $\beta G (\gamma +  \Phi'(\tQ_t))$, we continuously double the current estimation until an appropriate value is found (Steps~\ref{alg1:5}-\ref{alg1:7}). After that,  we set the learning rate 
\begin{equation}    \label{eq:eta}
     \eta_k = D ( 2 \tG_k T^{3/4} )^{-1},
\end{equation}
and construct the function 
\begin{equation}    \label{eq:F}
    F_{\ts_k:t}(\mathbf{x})= \eta_k \sum_{\tau=\ts_k}^{t} \left\langle\nabla \tf_{\tau}\left(\mathbf{x}_{\tau}\right), \mathbf{x}\right\rangle+\left\|\mathbf{x}-\mathbf{x}_{\ts_k}\right\|_{2}^{2} 
\end{equation}
based on the historical gradients $\nabla \tf_t(\x_t)$ since the round $\ts_k$ (Step~\ref{alg1:8}), where $\ts_k$ denotes the first round that utilizes the estimation $\Tilde{G}_k$. At the Step~\ref{alg1:9}, we compute $\v_t$ according to 
\begin{equation}    \label{eq:v}
    \mathbf{v}_{t} \in \underset{\mathbf{x} \in \mathcal{K}}{\operatorname{argmin}}\left\langle\nabla F_{\ts_k:t}\left(\mathbf{x}_{t}\right), \mathbf{x}\right\rangle, 
\end{equation}
and set the step size as 
\begin{equation}    \label{eq:sigma}
    \sigma_{\ts_k,t} = 2(t - \ts_k + 1)^{-1/2}.
\end{equation}
Finally, we update the decision $\x_{t+1}$ for the next round as shown below (Step~\ref{alg1:10}):
\begin{equation}    \label{eq:update:x}
    \x_{t+1}=\x_{t}+\sigma_{\ts_k,t}\left(\v_{t}-\x_{t}\right).
\end{equation}

\noindent
By choosing the exponential Lyapunov function $\Phi(x) = \exp  ( 2^{-1}T^{-3/4} x) - 1$, we establish the following theorem.
\begin{theorem}    \label{thm:cvx}
    Let $\beta = (2^6 GD)^{-1}$ and $\gamma = 1$. Under Assumptions~\ref{assump:K-bound}~and~\ref{assump:Lipschitz}, if the loss functions and the constraint functions are general convex, Algorithm~\ref{alg1} ensures the bounds of
    \begin{equation}
        \Reg_T =  \O(T^{3/4}),~Q_T = \O(T^{3/4} \log T). \nonumber
    \end{equation}
\end{theorem}
\noindent
\textbf{Remark.} 
Compared to the $\O(T^{3/4}\sqrt{\log T})$ regret bound and the $\O(T^{7/8})$ CCV bound in \citet{ICML:2024:Garber}, our results for both metrics are tighter. The underlying reasons can be attributed to: (i) the choice of projection-free methods. Under the drift-plus-penalty framework, \citet{ICML:2024:Garber} choose to run the projection-free LOO-BOGD method, which, due to its complex design, necessitates  additional effort to balance the costs of linear optimization and the performance. In contrast, our proposed method is inherently simpler, naturally requiring only one linear optimization per round; (ii) the specification of $\Phi(x)$. \citet{ICML:2024:Garber} implicitly choose $\Phi(x)=x$, which potentially fails to balance regret and CCV for general convex losses, leading to looser results. Furthermore, it should be emphasized that even if $\Phi(x)$ in \citet{ICML:2024:Garber} is replaced with the exponential function, the complex management of linear optimization costs in LOO-BOGD   still prevents it from yielding  the same results as ours.

\begin{algorithm}[t]
    \caption{Strongly Convex Variant of OFW  with Time-Varying Constraints (SCOFW-TVC)}
    \label{alg2} 
    \textbf{Input:} Hyper-parameters $\beta$, $\gamma$, and the function $\Phi(\cdot)$, and the modulus of strong  convexity $\alpha_f$
    \begin{algorithmic}[1]
        \State  Choose any $\x_1 \in \K$     \label{alg2:1}
        \For{$t = 1$ to $T$}
            \State  Play $\x_t$, and suffer $f_t(\x_t)$ and $g_t(\x_t)$ \label{alg2:3}
            \State Construct $Q_t$ and  $\tf_t(\x)$ according to \eqref{eq:Q} and \eqref{eq:tf} \label{alg2:4}
            \State Set $F_{t}^{sc}(\x)$ according to   \eqref{eq:str:F} \label{alg2:5}
            \State Compute $\v_t$ and $\sigma_{t}^{sc}$ according to \eqref{eq:str:v}  and  \eqref{eq:sigma:str:f}    \label{alg2:6}     
            \State Update $\x_{t+1}$ according to  \eqref{eq:update:str:x}  \label{alg2:8}
        \EndFor
    \end{algorithmic}
\end{algorithm}

\paragraph{Strongly Convex Losses.} In this case, note that for the $\alpha_f$-strong convex $f_t(\x)$, the surrogate loss $\tf_t(\x)$ defined in \eqref{eq:tf} is $\gamma \beta \alpha_f$-strongly convex. Therefore, we can employ the strongly convex variant of OFW to minimize $\tf_t(\x)$. In this paper, we choose the SCOFW method proposed by \citet{AAAI:2021:Wan}, because of its simplicity and  agnosticism to $\Tilde{G}$. The detailed procedures are given  in Algorithm~\ref{alg2}.

Specifically, we first choose any decision $\x_1 \in \K$ for initialization (Step~\ref{alg2:1}). Then,  at each round $t$, we make the decision $\x_t$, suffer the loss $f_t(\x_t)$ and the constraint $g_t(\x_t)$, and construct $Q_t$ and $\tf_t(\x)$ according to \eqref{eq:Q} and \eqref{eq:tf} (Steps~\ref{alg2:3}-\ref{alg2:4}). At Steps~\ref{alg2:5}-\ref{alg2:6}, we construct  $F_{t}^{sc}(\x)$ in the following way:
\begin{equation}    \label{eq:str:F}
    F_{t}^{sc}(\x)=  \sum\nolimits_{\tau=1}^{t} \left[ \left\langle\nabla \tf_{\tau}\left(\x_{\tau}\right), \x\right\rangle  + C_1 \|\x - \x_\tau\|_2^2 \right] 
\end{equation}
where we denote $C_1 = \gamma \beta  \alpha_f  / 2$ for brevity, and compute   $\v_t$ according to:  
\begin{equation}    \label{eq:str:v}
    \mathbf{v}_{t} \in \underset{\mathbf{x} \in \mathcal{K}}{\operatorname{argmin}}\left\langle\nabla F_{t}^{sc}\left(\mathbf{x}_{t}\right), \mathbf{x}\right\rangle  
\end{equation}
and $\sigma_{t}^{sc}$ according to
\begin{equation}    \label{eq:sigma:str:f}
        \sigma_{t}^{sc}=\underset{\sigma \in[0,1]}{\operatorname{argmin}}~ F_{t}^{sc}(\x_t + \sigma(\v_t - \x_t)).
\end{equation}
At the Step~\ref{alg2:8}, we update the decision $\x_{t+1}$ for the next round according to  
\begin{equation}    \label{eq:update:str:x}
    \x_{t+1}=\x_{t}+\sigma_{\ts_k,t}\left(\v_{t}-\x_{t}\right).
\end{equation}
By choosing the quadratic Lyapunov function $\Phi(x) = x^2 + x$, we establish the following theoretical results.

\begin{theorem}    \label{thm:str:f}
    Let $\beta =  G^{-1} D^{-1} T^{-2/3}$ and $\gamma =   G / (G + \alpha_f D)$.  Under Assumptions~\ref{assump:K-bound}~and~\ref{assump:Lipschitz}, if the loss functions  are $\alpha_f$-strongly convex, and the constraint functions are general convex,   Algorithm~\ref{alg2} ensures the   bounds of 
    \begin{equation}
        \Reg_T = \O(T^{2/3}),~Q_T =  \O(T^{5/6}). \nonumber
    \end{equation}
\end{theorem}
\noindent
\textbf{Remark.} Theorem~\ref{thm:str:f} provides the \textit{first} theoretical guarantees for the strongly convex losses in projection-free COCO, which are tighter than those  in \citet{ICML:2024:Garber} for general convex losses.

\subsection{Algorithms for Bandit Setting}
In this section, we investigate the bandit setting, where only the function value is available. To handle the more challenging setting, we introduce the one-point gradient estimator \citep{SODA:2005:Flaxman}, which can approximate the gradient with a single function value. 

\paragraph{One-Point Gradient Estimator.}
For a function $f(\bx)$, we define its $\delta$-smooth version as 
\begin{equation}
    \label{eq:smooth_version}
    \hat{f}_\delta(\mathbf{x})=\mathbb{E}_{\mathbf{u} \sim \mathcal{B}^d}[f(\mathbf{x}+\delta \mathbf{u})]
\end{equation}
which satisfies the following lemma \cite[Lemma 1]{SODA:2005:Flaxman}.
\begin{lemma}
    \label{lemma:one-point}
  Let $\delta > 0$, $\hat{f}_\delta(\mathbf{x})$ defined in \eqref{eq:smooth_version} ensures
  \begin{equation}
    \nabla \hat{f}_\delta(\mathbf{x})=\mathbb{E}_{\mathbf{u} \sim \mathcal{S}^d}\left[ (d/\delta) f(\mathbf{x}+\delta \mathbf{u}) \mathbf{u}\right]
  \end{equation}
  where  $\mathcal{S}^d$ denotes  the unit sphere  in $\mathbb{R}^d$.
\end{lemma}

\noindent
To exploit the one-point gradient estimator, we define the shrunk set of $\K$ as stated below
$$
    \mathcal{K}_{\delta} = (1 - \delta / r) \mathcal{K} = \{(1-\delta / r) \mathbf{x} \mid \mathbf{x} \in \mathcal{K} \},
$$
where $0 < \delta < r$ denotes the shrunk parameter.

Compared to our methods   for the full-information setting, we make the following modifications:
\begin{itemize}
    \item At each round $t$, the decision $\x_t$ consists of two parts:
        \begin{equation}
            \label{eq:bdt:x}
            \x_t=\y_t + \delta \u_t
        \end{equation}
        where $\y_t \in \K_{\delta}$ denotes an auxiliary decision learned from historical information, and $\u_t \sim \mathcal{S}^{d}$ is uniformly sampled from $\mathcal{S}^{d}$;

    \item The gradient of $\tf_t(\x_t)$ is approximated by the one-point gradient estimator:
        \begin{equation}
            \label{eq:tnab}
            \tnabla_{t} = (d/\delta) [\tf_t(\x_t)] \mathbf{u}_{t},
        \end{equation}
    so that we can adhere to the update rules in our previous methods;

    \item To manage the approximate error introduced by \eqref{eq:tnab}, we employ the blocking technique \citep{AISTATS:2020:Garber,COLT:2020:Hazan} for decision updates, i.e.,~dividing the time horizon $T$ into equally-sized blocks and only updating decisions at the end of each block.
\end{itemize}
In the bandit setting, we also investigate the general convex losses and the strongly convex losses.

\begin{algorithm}[t]
    \caption{Bandit Frank-Wolfe with Time-Varying Constraints (BFW-TVC)}
    \label{alg:bandit:cvx} 
    \textbf{Input:} Hyper-parameters $\beta$, $\gamma$, $c$, $K$, $\epsilon$, and the function $\Phi(\cdot)$ 
    \begin{algorithmic}[1]
        
        \State Choose any $ \hat{\y}_1 \in \K_{\delta_1}$, and set $ \Tilde{G}_1 =  m = 1$ \label{alg3:1}
        
        \For{$t = 1$ to $T$}
            \State Set $\y_t = \hat{\y}_m$, and  play $\x_t$ according to \eqref{eq:bdt:x}    \label{alg3:2}
            
            \State Suffer $f_t(\x_t)$ and $g_t(\x_t)$    \label{alg3:3}
            
            \State Construct $Q_t$ and  $\tf_t(\x)$ according to \eqref{eq:Q} and \eqref{eq:tf}  \label{alg3:4}
            
            \State Compute $\tnabla_t$ according to \eqref{eq:tnab}  \label{alg3:5}

            \If{$t$ mod $K = 0$}
                \While{ $\tG_k < \beta G (\gamma +      \Phi'(\tQ_\tau))$,$\forall \tau$ in block}       \label{alg3:7}
                    
                    \State Set $\tG_{k+1} = 2\tG_k$,  $k = k+1$   
                
                \EndWhile    \label{alg3:9}
                
                \State Compute $\hnabla_{m} = \sum_{\tau =t-K+1}^t \tnabla_\tau$      \label{alg3:10}

                \State Set $\eta_k$ and $F_{b_k:m}(\y)$  according to \eqref{eq:bdt:eta} and \eqref{eq:bdt:F} \label{alg3:11}

                \State Set $\ty_1 = \hy_m$ and $\tau = 0$  \label{alg3:12}

                \Repeat    \label{alg3:13}
                    \State Set   $\tau = \tau + 1$    \label{alg3:14} 

                    \State Update   $\v_\tau$ and  $\sigma_{\tau}$ according to \eqref{eq:update:bdt:v} and \eqref{eq:bdt:sigma:f} 
                    
                    % % \State Calculate  $\sigma_{\tau}$  according to    \eqref{eq:bdt:sigma:f} 

                    \State Compute $\ty_{\tau+1}$ according to  \eqref{eq:bdt:update:y} 
                \Until{$\langle \nabla F_{b_k: m}(\ty_{\tau}), \ty_\tau - \v_\tau  \rangle \leq \epsilon$} \label{alg3:18}

                \State Set $\hy_{m+1} = \ty_{\tau+1}$ and $m = m + 1$ 
            \EndIf
        \EndFor
    \end{algorithmic}
\end{algorithm}

\paragraph{General Convex Losses.} 
In this case, we incorporate the modifications~\eqref{eq:bdt:x}~and~\eqref{eq:tnab}, and the blocking technique into Algorithm~\ref{alg1}. The detailed procedures are summarized in Algorithm~\ref{alg:bandit:cvx}. Specifically, for initialization, we set $\Tilde{G}_1 = m = 1$, and choose any $\hy_1 \in \K_{\delta}$ (Step~\ref{alg3:1}). At each round $t$, we update $\y_t = \hat{\y}_m$ where $\hy_{m} \in \K_{\delta}$ is the auxiliary decision used in the block $m$, make the decision $\x_t$ according to \eqref{eq:bdt:x}, and suffer $f_t(\x_t)$ and $g_t(\x_t)$ (Steps~\ref{alg3:2}-\ref{alg3:3}). Then, we construct the CCV $Q_t$, the surrogate loss function $\tf_t(\x)$ and the gradient estimation  $\tnabla_t$ according to \eqref{eq:Q}, \eqref{eq:tf} and \eqref{eq:tnab}, respectively (Steps~\ref{alg3:4}-\ref{alg3:5}). At the end of the block $m$, we update our decision. To be precise, we first evaluate the current  guess  $\Tilde{G}_k$ for the gradient norm bound: if it is unsuitable, we double the value  until  an appropriate $\Tilde{G}_k$ is found (Steps~\ref{alg3:7}-\ref{alg3:9}). At the Step~\ref{alg3:10}, we compute the   cumulative gradient estimation $\hnabla_{m} = \sum_{\tau =t-K+1}^t \tnabla_\tau$  where $K$ denotes the block size. At the Step~\ref{alg3:11}, we set $\eta_k$ according to 
\begin{equation}    \label{eq:bdt:eta}
     \eta_k =  cD ( d M \tG_k T^{3/4} )^{-1},
\end{equation}
and construct $F_{b_k:m}$ according to 
\begin{equation}
    \label{eq:bdt:F}
    F_{b_k:m}(\mathbf{y})=\eta_k  \sum\nolimits_{\tau=b_k}^m\left\langle \hnabla_\tau, \mathbf{y}\right\rangle+\left\|\mathbf{y}-\mathbf{y}_{\ts_k}\right\|^2_{2},
\end{equation}
where $b_k$ denotes the first block that utilizes the estimation $\Tilde{G}_k$, and $\ts_k$ denotes the first round of $b_k$. Next, we update the auxiliary decision for the next block, and set $\ty_1 = \y_m$ and $\tau=0$ (Step~\ref{alg3:11}). At   Steps~\ref{alg3:13}-\ref{alg3:18}, we repeat the following procedures: updating $\tau = \tau + 1$, computing $\v_\tau$ according to
\begin{equation}    \label{eq:update:bdt:v}
    \mathbf{v}_{\tau} \in \underset{\y \in \mathcal{K}_\delta}{\operatorname{argmin}}\left\langle\nabla F_{b_k:m}\left(\ty_{\tau}\right), \y \right\rangle,
\end{equation}
and $\sigma_{\tau}$ according to
\begin{equation}    \label{eq:bdt:sigma:f}
    \sigma_{\tau} =\argmin_{\sigma \in[0,1]} F_{b_k:m} \left(\ty_{\tau}+\sigma \left(\v_\tau -\ty_{\tau}\right)\right), 
\end{equation} 
and updating $\ty_{\tau+1}$ according to 
\begin{equation}    \label{eq:bdt:update:y}
    \ty_{\tau+1}=\ty_{\tau}+\sigma_{\tau}\left(\v_{\tau}-\ty_{\tau}\right),
\end{equation}
until the stop condition  $\langle \nabla F_{b_k: m}(\ty_{\tau}), \ty_\tau - \v_\tau  \rangle \leq \epsilon$ is satisfied. After that, we set the auxiliary decision $\hy_{m+1} = \ty_{\tau+1}$ for the next block.

With the configurations of the exponential Lyapunov function  $\Phi(x) = \exp  ( 2^{-1}T^{-3/4} x) - 1$ and suitable parameters, we obtain the following theorem.
\begin{theorem}    \label{thm:bdt:cvx}
    Let  $\gamma = 1$,  $K = T^{1/2}$, $\epsilon = 4D^2T^{-1/2}$ and $c > 0$ be a constant satisfying $\delta = cT^{-1/4} \leq r$, and $\beta = C_2^{-1}$ where $C_2 =    2^4  G   (  cD/r + 3c + 1 +  2cD /(dM)  +  dMD/c  )$. Under Assumptions~\ref{assump:K-bound}~and~\ref{assump:Lipschitz}, if the loss functions and the constraint functions are general convex,  Algorithm~\ref{alg:bandit:cvx} ensures the   bounds of 
    \begin{equation}
        \E[\Reg_T] = \O(T^{3/4}),~Q_T = \O(T^{3/4} \log T). \nonumber
    \end{equation}
\end{theorem}

\noindent
\textbf{Remark.} Theorem~\ref{thm:bdt:cvx} presents tighter regret and CCV bounds, compared to the $\O(T^{3/4} \sqrt{\log T})$ regret bound and the $\O(T^{7/8} \log T  )$ CCV bound in \citet{ICML:2024:Garber}.

\begin{algorithm}[t]
    \caption{Strongly Convex Variant of BFW  with Time-Varying Constraints (SCBFW-TVC)}
    \label{alg:bandits:str} 
    \textbf{Input:} Hyper-parameters $\beta$, $\gamma$, $\delta$, $K$, $L$, and the function $\Phi(\cdot)$, and the modulus of strong  convexity $\alpha_f$
    \begin{algorithmic}[1]
        \State  Choose any $\hy_1 \in \K_\delta$, and set $m=1$ 
        \For{$t = 1$ to $T$}
             \State Set $\y_t = \hat{\y}_m$, and  play $\x_t$ according to \eqref{eq:bdt:x}    \label{alg4:2}
            
            \State Suffer $f_t(\x_t)$ and $g_t(\x_t)$    \label{alg4:3}
                 
            \State Construct $Q_t$ and  $\tf_t(\x)$ according to \eqref{eq:Q} and \eqref{eq:tf}   \label{alg4:4}

            \State Compute  $\tnabla_t$ according to \eqref{eq:tnab}    \label{alg4:5}

            \If{$t$ mod $K = 0$}
                \State Compute $\hnabla_{m} = \sum_{\tau =t-K+1}^t \tnabla_\tau$      \label{alg4:7}

                \State Set $F_m^{sc}(\y)$  according to \eqref{eq:bdt:str:F}, and $\ty_1 = \y_m$   \label{alg4:8}

                \For {$\tau = 1$ to $L$}    \label{alg4:9}

                    \State Compute $\v_\tau^{sc}$ according to \eqref{eq:bdt:v:str}
            
                    % \State Update $\v_\tau^{sc}$ and $\sigma_{t}^{sc}$ according to \eqref{eq:bdt:v:str} and \eqref{eq:bdt:sigma:str:f}
                    
                    \State Calculate  $\sigma_{t}^{sc}$ according to \eqref{eq:bdt:sigma:str:f}

                    \State Update $\ty_{\tau+1}$ according to \eqref{eq:bdt:update:y:str}
                    
                \EndFor    \label{alg4:13}
                
                \State Set $\hy_{m+1} = \ty_{L+1}$ and $m = m + 1$ \label{alg4:14}
            \EndIf
            
        \EndFor
    \end{algorithmic}
\end{algorithm}

\paragraph{Strongly Convex Losses.} 
In this case, we also  employ the one-point gradient estimator and  the blocking technique, and  summarize the procedures in Algorithm~\ref{alg:bandits:str}. Overall, our method for strongly convex losses is similar to Algorithm~\ref{alg:bandit:cvx}, with the primary difference in the update of  decisions. Specifically,  at the end of each block $m$, we first compute the cumulative gradient estimation    $\hnabla_{m} = \sum_{\tau =t-K+1}^t \tnabla_\tau$ (Step~\ref{alg4:7}), and then construct $F_{m}^{sc}(\y)$ as shown below:
\begin{equation}    \label{eq:bdt:str:F}
    F_{m}^{sc}(\y)=  \sum\nolimits_{\tau=1}^{m}   \left\langle\hnabla_\tau, \y\right\rangle  + C_3 \|\y\|_2^2
\end{equation}
where we denote $C_3 = \gamma \beta  \alpha_f t / 2$ for brevity, and set $\ty_1 = \y_m$ (Step~\ref{alg4:8}).  Next, we repeat the following procedures for $L$ times to refine the auxiliary  decision (Steps~\ref{alg4:9}-\ref{alg4:13}): at the iteration $\tau \in [L]$,
computing $\v_\tau^{sc}$ according to 
\begin{equation}    \label{eq:bdt:v:str}
    \v_\tau^{sc} \in \underset{\mathbf{y} \in \mathcal{K}_\delta}{\operatorname{argmin}}\left\langle\nabla F_{m}^{sc}\left(\ty_{\tau}\right), \mathbf{y}\right\rangle,
\end{equation}
calculating $\sigma_{t}^{sc}$  according to
\begin{equation}    \label{eq:bdt:sigma:str:f}
    \sigma_{\tau}^{sc}=\argmin_{\sigma \in[0,1]} F_{m}^{sc}\left(\ty_{\tau}+\sigma \left(\v_\tau^{sc}-\ty_{\tau}\right)\right), 
\end{equation}
and updating $\ty_{\tau+1}$ according to 
\begin{equation}    
    \label{eq:bdt:update:y:str}
    \ty_{\tau+1}=\ty_{\tau}+\sigma_{\tau}^{sc}\left(\v_\tau^{sc}-\ty_{\tau}\right).
\end{equation}
Finally, we set the  auxiliary decision for the next block as $\y_{m+1} = \ty_{L+1}$ (Step~\ref{alg4:14}).

By setting the quadratic Lyapunov function $\Phi(x) = x^2$ and proper parameters, we obtain the following theorem.

\begin{theorem}    \label{thm:bdt:str:f}
    Let $\beta =  G^{-1} D^{-1} T^{-2/3}$, $\gamma =   G / (G + \alpha_f D)$,  $K=L=T^{2/3}$, and $\delta = c T^{-1/3}$ with $c > 0$ satisfying $cT^{-1/3} < r$, and $\gamma = \O (T^{2 / 3})$.  Under Assumptions~\ref{assump:K-bound}~and~\ref{assump:Lipschitz}, if the loss functions  are $\alpha_f$-strongly convex, and the constraint functions are general convex,   Algorithm~\ref{alg:bandits:str} ensures the   bounds of 
    \begin{equation}
         \E\left[\Reg_T\right]  =  \O(T^{2/3}  \log T),~Q_T =  \O(T^{5/6}  \log T).\nonumber
    \end{equation}
\end{theorem}
\noindent
\textbf{Remark.} Theorem~\ref{thm:bdt:str:f} provides the \textit{first} regret and CCV bounds for the strongly convex case with bandit feedback in projection-free COCO. By utilizing the strong convexity of $f_t(\cdot)$, both of our results are tighter than  those established for the general convex losses in \citet{ICML:2024:Garber}.

\begin{figure}[t]
    \centering
        \begin{subfigure}{0.23\textwidth}
            \includegraphics[width=\linewidth, height=1.35in]{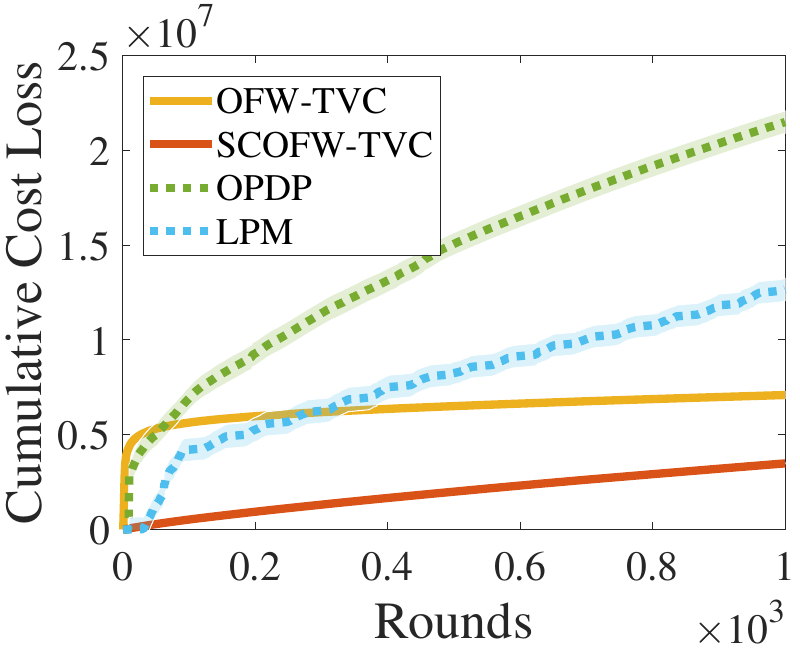}  
            \caption{Cumulative Cost Loss}
        \end{subfigure} 
        \begin{subfigure}{0.23\textwidth}
            \includegraphics[width=\linewidth, height=1.35in]{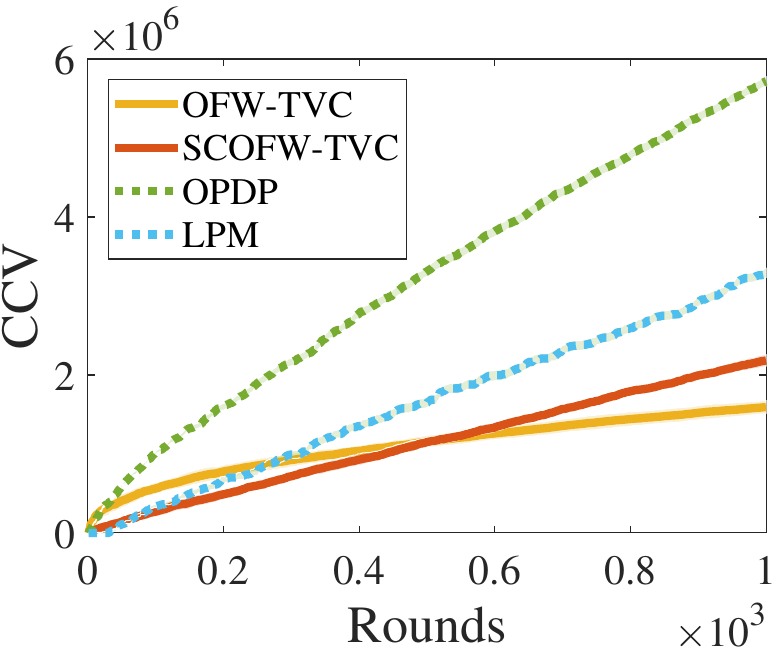}
            \caption{CCV}
        \end{subfigure}

        \caption{ Experimental results on the MovieLens dataset.}
                
        \label{fig:1}
\end{figure}

\section{Experiments}
In this section, we conduct empirical studies on real-world datasets to evaluate  our theoretical findings.

\paragraph{General Setup.} 
We investigate the online matrix completion problem \citep{ICML:2012:Hazan,ArXiv:2023:Lee}, the goal of which is to generate a matrix $X$  in an online manner to approximate the target matrix $M \in \mathbb{R}^{m\times n}$. Specifically, at each round $t$, the learner receives a sampled data $(i, j)$ with the value $M_{ij}$ from the observed subset $O$ of $M$. Then, the learner chooses a matrix $X$ from the trace norm ball $\K = \{ X |  \|X\|_* \leq \delta, X \in \mathbb{R}^{m \times n}\}$  where $\delta > 0$ is the parameter, and suffers the strongly convex cost loss $f_t(X_t) =  \sum_{(i, j) \in O} (X_{ij} - M_{ij} )^2/2$ and the constraint loss $g_t(X_t) = \text{Tr}(P_t X_t)$ where $P_t$ is uniformly sampled from $[-1, 1]^{n \times m}$. The experiments are conducted with $\delta=10^4$ on two real-world datasets: MovieLens\footnote{https://grouplens.org/datasets/movielens/100k/} for the full-information setting, and Film Trust \cite{IJCAI:2013:Guo} for the bandit setting.

\paragraph{Baselines.} We choose three projection-free COCO methods as the contenders: (i) OPDP \cite[Algorithm 1]{ArXiv:2023:Lee} and LPM \cite[Algorithm 4]{ICML:2024:Garber} for the full-information setting; (ii) LBPM \cite[Algorithm 5]{ICML:2024:Garber} for the bandit setting. All parameters of each method are set according to their theoretical suggestions, and we choose the best hyper-parameters from the range of $[10^{-5}, 10^{-4}, \dots, 10^{4}, 10^{5}]$.

\begin{figure}[t]
    \centering
        \begin{subfigure}{0.23\textwidth}
            \includegraphics[width=\linewidth, height=1.35in]{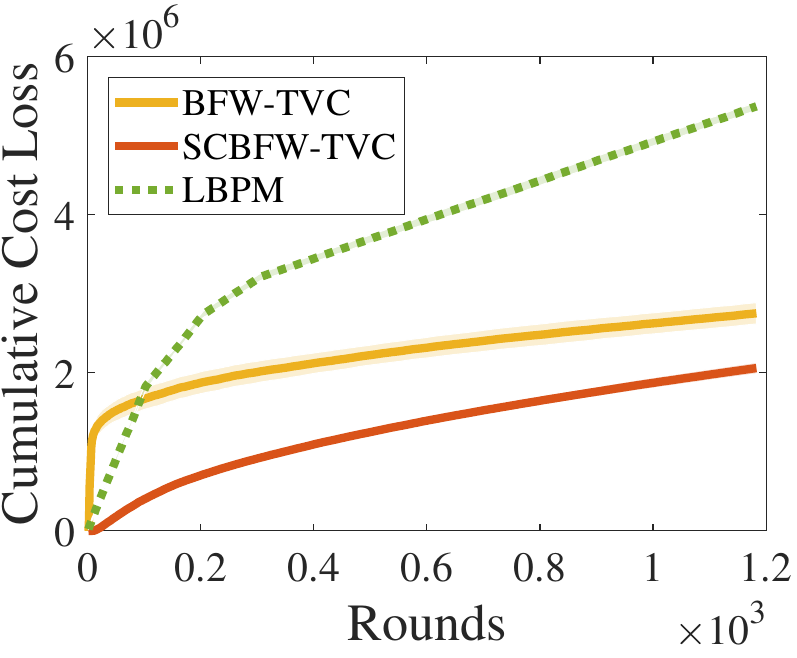}  
            \caption{Cumulative Cost Loss}
        \end{subfigure} 
        \begin{subfigure}{0.23\textwidth}
            \includegraphics[width=\linewidth, height=1.35in]{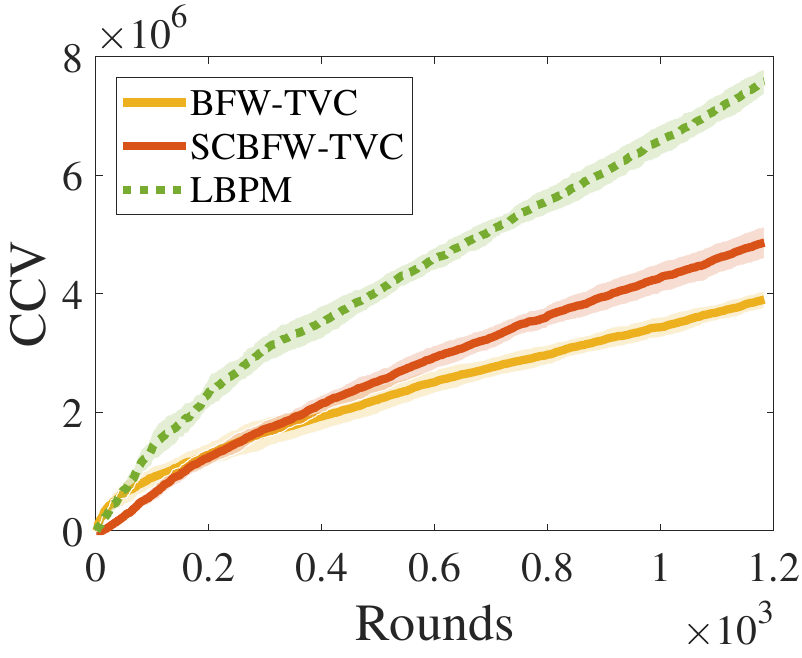}
            \caption{CCV}
        \end{subfigure}

        \caption{ Experimental results on the Film Trust dataset.}
                
        \label{fig:2}
\end{figure}

\paragraph{Results.} All experiments are repeated $10$ times and we report experimental results (mean and standard deviation) in Figures~\ref{fig:1}~and~\ref{fig:2}. As evident from the results, in the full-information setting,  OFW-TVC outperforms its competitors significantly  in terms of  both two metrics.  Moreover, by utilizing the strong convexity of $f_t(\cdot)$, our SCOFW-TVC yields the lowest cumulative cost loss, albeit with a slight compromise on CCV. Similarly, in the bandit setting, it can be observed that our methods consistently outperform others, aligning with the theoretical guarantees.

\section{Conclusion and Future Work}
In this paper, we investigate projection-free COCO and propose a series of methods for the full-information and bandit settings. The key idea is to utilize the Lyapunov-based technique to construct a composite surrogate loss, consisting of the original cost   and the constraint loss, and employ parameter-free variants of OFW running over the surrogate loss to simultaneously optimize the regret and CCV. In this way, we improve previous results for general convex cost losses and establish \textit{novel} regret and CCV bounds for strongly convex cost losses. During the analysis, we   propose the \textit{first} parameter-free variant of OFW for general convex losses, which may hold independent interest. Finally, empirical studies have verified our theoretical findings.

Currently, for   strongly convex losses, we improve the regret bound from $\O(T^{3/4})$ to $\O(T^{2/3})$ , but sacrifice another metric CCV with a marginally  looser bound of $\O(T^{5/6})$, compared to our results for general convex losses. This phenomenon may be due to the  potential  impropriety of the quadratic Lyapunov function. Hence, one possible solution is to choose other more powerful functions, which seems highly non-trivial, and we leave it as future work.

\section{Acknowledgments}
This work was partially supported by National Science and Technology Major Project (2022ZD0114801), and NSFC (U23A20382, 62306275). We would like to thank the anonymous reviewers for their constructive suggestions.

\bibliography{aaai25}

\appendix
\onecolumn

\section{Theoretical Analysis}

\subsection{Proof of Theorem~\ref{thm:cvx}}
We first introduce the following lemma that reveals the relationship $\Reg_T$ and $\Reg_T'$.
\begin{lemma}    \label{lem:regret}
    Let $\tf_t(\cdot)$ be defined in \eqref{eq:tf}, and $\Reg_T'$ denote the regret in terms of $\tf_t(\cdot)$. Then, we have 
    \begin{equation}    \label{eq:cvx:regret}
    \begin{split}
        \Reg_T \leq \frac{1}{\gamma \beta}\left(\Reg'_T - \Phi(\tQ_T)\right).
    \end{split} 
\end{equation}
\end{lemma}

Now, we focus on $\Reg_T'$. By utilizing the convexity of $\tf_t(\cdot)$, we have
\begin{equation}
\begin{split}
    \Reg'_T = & \sum_{t=1}^T \left[\tf_t(\x_t) - \tf_t(\x^*) \right] \leq \sum_{t=1}^T \langle \nabla  \tf_t(\x_t), \x_t - \x^* \rangle \\
     \leq & \sum_{k=1}^K \underbrace{ \left[ \sum_{j=1}^{t_k} \langle \nabla  \tf_t(\x_t), \x_t - \x_{\ts_k:t}^* \right. \rangle}_\ta +  \underbrace{ \left. \sum_{j=1}^{t_k}  \langle \nabla  \tf_t(\x_{t}), \x_{\ts_k:t}^* - \x^* \rangle  \right] }_\tb,    \nonumber
\end{split}
\end{equation}
where $t_k$ denotes the size of the $k$-th block that employ   $\tG_k = 2^{k-1} G_1 = 2^{k-1}$,  and  $\ts_k = \sum_{i=1}^{k-1} t_i + 1$ denote the first round of block $k$, and  $t = \ts_k - 1 + j$ and $\x_{\ts_k:t}^* = \argmin_{\x \in \K} F_{\ts_k:t-1}(\x)$.
Then, we analyze the regret on $k$-th block, which consists of two terms.  To upper bound $\ta$, we first introduce the following lemma.
\begin{lemma}    \label{lem:cvx:F}
    Let $s$ be the first round in the block $k$ and  $\sigma_{\ts_k,t} =2 (t -  \ts_k + 1)^{-1/2}$. By setting $\eta_k = \frac{D}{2  \tG_k  T^{3/4} }$, we have
    \begin{equation}
        F_{\ts_k:t-1}(\x_t) - F_{\ts_k:t-1}(\x_{\ts_k:t}^*) \leq 2D^2 \sigma_{\ts_k,t}.  \nonumber
    \end{equation}
\end{lemma}
\noindent
By applying Lemma~\ref{lem:cvx:F}, we have
\begin{equation}
\begin{split}    \label{eq:cvx:ta}
    \ta \leq &  \sum_{j=1}^{t_k} \| \nabla \tf_t(\x_t) \|_2 \| \x_t - \x_{\ts_k:t}^* \|_2 
    \overset{\eqref{eq:property:str}}{\leq}    \sum_{j=1}^{t_k} \| \nabla \tf_t(\x_t) \|_2 \sqrt{F_{\ts_k:t-1}(\x_t) - F_{\ts_k:t-1}(\x_{\ts_k:t}^*)} \\
    \leq &  \tG_k \sum_{j=1}^{t_k}   \sqrt{F_{\ts_k:t-1}(\x_t) - F_{\ts_k:t-1}(\x_{\ts_k:t}^*)}  
    \leq  \tG_k D  \sum_{j=1}^{t_k}  \sqrt{2 \sigma_{\ts_k,t}}  
    \leq   4 \tG_k D t_k^{3/4} = 2^{k+1} D t_k^{3/4},
\end{split}
\end{equation}
where the penultimate inequality is due to the fact that $t = \ts_k - 1 + j$ and $\sigma_{\ts_k,t} = 2 (t - \ts_k + 1)^{-1/2}  = 2 j^{-1/2}$, and the last inequality is due to  $\tG_k = 2^{k-1} G_1 = 2^{k-1}$ for the $k$-th block.
\noindent
For $\tb$, we introduce the following lemma.
\begin{lemma}    \label{lem:cvx:tb}
    Let $\x_{\ts_k:t}^* = \argmin_{\x \in \K} F_{\ts_k:t-1}(\x)$. Then,  we have
    \begin{equation}
        \sum_{j=1}^{t_k}  \langle \nabla  \tf_t(\x_t), \x_{\ts_k:t}^* - \x^* \rangle  \leq \frac{D^2}{\eta_k} + \eta_k \sum_{j=1}^{t_k} \| \nabla \tf_t(\x_t) \|_2^2.  \nonumber
    \end{equation}
\end{lemma}

\noindent
Substituting the setting of $\eta_k = \frac{D}{2 \tG_k T^{3/4} }$ into Lemma~\ref{lem:cvx:tb}, we obtain
\begin{equation}    \label{eq:cvx:tb}
\tb \leq \frac{D^2}{\eta_k} + \eta_k \sum_{j=1}^{t_k} \| \nabla \tf_t(\x_t) \|_2^2 \leq 2 \tG_k D (T^{3/4} + T^{1/4}) \leq 2^{k+1} D T^{3/4}.
\end{equation}
Combining \eqref{eq:cvx:ta} and \eqref{eq:cvx:tb}, we have
\begin{equation} 
    \begin{split}
        \Reg'_T \leq &  D \sum_{k=1}^K 2^{k+1} t_k^{3/4} + D T^{3/4} \sum_{k=1}^K 2^{k+1} \\
                \leq & D \left(\sum_{k=1}^K (t_k^{3/4})^{4/3}\right)^{3/4}\left(\sum_{k=1}^K (2^{k+1})^{4}\right)^{1/4} + 2^{K+3} D T^{3/4} 
                \leq  2^{K+4} D T^{3/4}    \nonumber
    \end{split}    
\end{equation}
where the second step is due to the Hölder's inequality and the last step is due to the fact that $\sum_{k=1}^K t_k = T$. Recall that in the last block $K$, we have 
\begin{equation}
    2^{K-2} = \tG_{K-1} \leq \beta G (\gamma +  \Phi'(\tQ_T)) \leq \tG_K = 2^{K-1},     \nonumber
\end{equation}
which implies that $2^{K+4} \leq 2^6 \beta G (\gamma +  \Phi'(\tQ_T))$ so that 
\begin{equation}    \label{eq:cvx:regret:prim}
     \Reg'_T \leq 2^6 \beta G D (\gamma +  \Phi'(\tQ_T))  T^{3/4}.
\end{equation}
Therefore, substituting \eqref{eq:cvx:regret:prim} into \eqref{eq:cvx:regret}   obtains
\begin{equation}
    \begin{split}
        \Reg_T = \sumT \left[\hf_t(\x_t) - \hf_t(\x_t) \right]
        \leq & \frac{1}{\gamma \beta}\left(2^6 \beta GD (\gamma +  \Phi'(\tQ_T)) T^{3/4} - \Phi(\tQ_T)\right) \\
        \leq &  \frac{1}{\gamma \beta}\left(2^6 \beta GD (\gamma +  \lambda \exp  ( \lambda \tQ_T )) T^{3/4} - \exp  ( \lambda \tQ_T  )\right) + \frac{1}{\gamma \beta} \\
        \leq & 2^6  GD T^{3/4} + \frac{\exp( \lambda \tQ_T )}{\gamma \beta} \left( 2^6 \beta GD \lambda T^{3/4} - 1\right)+ \frac{1}{\gamma \beta}
    \end{split}
     \nonumber
\end{equation}
where the second step is due to the choice of $\Phi(\tQ_t) = \exp  ( \lambda \tQ_t  ) - 1 $ and $\Phi'(\tQ_t) = \lambda \exp  ( \lambda \tQ_t  )$ for any $t \in [T]$.

\noindent
By setting $\beta = (2^6 GD)^{-1}$, $\lambda \leq T^{-3/4}$ and $\gamma = 1$, we have
\begin{equation}
    \Reg_T \leq  2^6   GD (T^{3/4} + 1).  \nonumber
\end{equation}
 
Finally, we specify the upper bound of CCV. From \eqref{eq:regret_trans}, we have the following relationship:
\begin{equation}      \label{eq:cvx:cvv}
    \begin{split}
        \Phi(\tQ_T) \leq  \Reg'_T -  \gamma  \beta \Reg_T \leq \Reg'_T +  \gamma  \beta GD T,
    \end{split} 
\end{equation}
where the last step is due to the $G$-Lipschitzness of $f_t(\cdot)$, i.e., for any $\x, \y \in \K$
\begin{equation}
    \left|f_{t}(\x)-f_{t}(\y)\right| \leq G\|\x-\y\|_2 \leq GD.  \nonumber
\end{equation}
Substituting \eqref{eq:cvx:regret:prim} into \eqref{eq:cvx:cvv}, we have
\begin{equation}    \label{eq:cvx:cvv:2}
    \begin{split}
        \Phi(\tQ_T) \leq & 2^6 \beta GD (\gamma +  \Phi'(\tQ_T)) T^{3/4}  +  \gamma  \beta GD T 
        \leq   \Phi'(\tQ_T) T^{3/4} + 2T.
    \end{split} 
\end{equation}
where the last step is due to the configurations of  $\beta = (2^6 GD)^{-1}$ and $\gamma = 1$. 
By setting $\Phi(\tQ_t) = \exp  ( \lambda \tQ_t  ) - 1 $ for any $t \in [T]$, we have 
\begin{equation}
    \begin{split}
        \exp(\lambda \tQ_T) - 1 \leq \lambda \exp(\lambda \tQ_T)  T^{3/4} + 2T.
    \end{split}
    \nonumber
\end{equation}
Rearranging the above inequality, we have
\begin{equation}
Q_T \leq \frac{1}{\lambda \beta} \ln \left( \frac{1 +    2T}{1 - \lambda T^{\frac{3}{4}}} \right) \leq 2^7 GD T^{3/4} \ln(2 + 4T),  \nonumber
\end{equation}
where the last step is due to $\lambda = 2^{-1}T^{-3/4}$.

\subsection{Proof of Theorem~\ref{thm:str:f}}
Similar to the proof of Theorem~\ref{thm:cvx}, the $\Reg_T$ is bound by 
\begin{equation}       \label{eq:str:f:1}
    \begin{split}
        \Reg_T \leq \frac{1}{\gamma \beta}\left(\Reg'_T - \Phi(\tQ_T)\right).
    \end{split} 
\end{equation}
Now, we focus on $\Reg_T'$ and introduce the following lemma \cite[Theorem3]{AAAI:2021:Wan}.
\begin{lemma}
    \label{lem:str:f}
    Let $\{h_t(\x)\}_{t=1}^T$ be a sequence of $\lambda$-strongly convex loss functions, and $G'$-Lipschitz over $\K$. Then, Algorithm~\ref{alg2} ensures
    \begin{equation}
        \sumT h_t(\x_t) - \min_{\x \in \K} \sumT h_t(x) \leq \frac{3 \sqrt{2} C T^{2 / 3}}{8}+\frac{C \log T}{8}+G' D    \nonumber
    \end{equation}
    where $C = 16(G' + \lambda D)^2 / \lambda$.
\end{lemma}
\noindent
Then, we apply Lemma~\ref{lem:str:f} over the loss functions $\{\tf_t(\x)\}_{t=1}^T$. Note that  given $f_t(\x)$ is $\alpha_f$-strongly convex, the function $\tf_t(\x)$ is  $\gamma \beta \alpha_f$-strongly convex. Therefore, by applying Lemma~\ref{lem:str:f}, we have 
\begin{equation}
\begin{split}
    \Reg'_T \leq & \frac{6 \sqrt{2} (G' + \gamma \beta \alpha_f D)^2  }{\gamma \beta \alpha_f} T^{2 / 3} + \frac{2 (G' + \gamma \beta \alpha_f D)^2 }{\gamma \beta \alpha_f} \log T + G' D \\
    \leq & \frac{14 (G' + \gamma \beta \alpha_f D)^2  }{\gamma \beta \alpha_f} T^{2 / 3} + G' D \\
    = & \frac{14 \beta (  \gamma (G + \alpha_f D ) +  G \Phi'(\tQ_T)    )^2  }{\gamma  \alpha_f}  T^{2 / 3} + \beta G D (\gamma +  \Phi'(\tQ_T)) \\
    = & \frac{14 \beta G^2 (  1 +  \Phi'(\tQ_T)    )^2 }{\gamma  \alpha_f} T^{2 / 3}  + \beta G D (\gamma +  \Phi'(\tQ_T))\\
    \leq & \frac{28 \beta G^2 }{\gamma  \alpha_f} T^{2 / 3}  (  1 +  \Phi'(\tQ_T)^2 )+ \beta G D (\gamma +  \Phi'(\tQ_T))
    \nonumber
\end{split}
\end{equation}
where the third step is due to $G' =  \beta G (\gamma +  \Phi'(\tQ_T))$ and the fourth step is due to $\gamma = G / (G + \alpha_f D)$, and the last step is due to the fact that $(a + b)^2 \leq 2(a^2 + b^2)$ for $\forall a, b \in \mathbb{R}$.

We set the function $\Phi(x) = x^2 + x$ with $\Phi'(x) = 2x+1$, and hence, $\Reg_T'$ is bounded by
\begin{equation}    \label{eq:str:f:2}
    \begin{split}
        \Reg'_T \leq &  \frac{28 \beta G^2 }{\gamma  \alpha_f} T^{2 / 3}  (  1 +   (2\tQ_T + 1)^2 )+ \beta G D (\gamma +  2\tQ_T + 1) \\ 
             \leq  & \frac{28 \beta G^2 }{\gamma  \alpha_f} T^{2 / 3}  (  3 +   8 \beta^2 Q_T^2)+ \beta G D (\gamma +  2\tQ_T + 1) \\
             \leq & \frac{224 \beta G^2 }{\gamma  \alpha_f} T^{2 / 3}   (1 + \beta^2 Q_T^2) + \beta G D (\gamma +  2\tQ_T + 1),
    \end{split}    
\end{equation}
where the second step is also due to $(a + b)^2 \leq 2(a^2 + b^2)$ for $\forall a, b \in \mathbb{R}$.

Substituting \eqref{eq:str:f:2} into \eqref{eq:str:f:1}, we obtain 
\begin{equation}
    \begin{split}
        \Reg_T \leq &  \frac{1}{\gamma \beta}\left( \frac{224 \beta G^2 }{\gamma  \alpha_f} T^{2 / 3}   (1 + \beta^2 Q_T^2) + \beta G D (\gamma +  2\tQ_T + 1) - \beta^2 Q_T^2 - \tQ_T\right) \\
        = & \frac{224  G^2 }{\gamma^2  \alpha_f} T^{2 / 3}   + \frac{GD(\gamma + 1)}{\gamma} + \left[\frac{224  G^2 }{\gamma  \alpha_f} T^{2 / 3} - \frac{1}{ \beta}\right] \frac{ \beta^2 Q_T^2}{\gamma}  + \left[2GD - \frac{1}{\beta} \right]  \frac{ \tQ_T }{\gamma} \\
       = &  \frac{224  G^2 }{\gamma^2  \alpha_f} T^{2 / 3}  + \frac{GD(\gamma + 1)}{\gamma}, \nonumber
    \end{split}
\end{equation}
where the last step is due to 
\begin{equation}
    \frac{224  G^2 }{\gamma  \alpha_f} T^{2 / 3} - \frac{1}{ \beta} \leq 0 \quad\text{and}\quad 2GD - \frac{1}{\beta} \leq 0
\end{equation}
with the setting of 
\begin{equation}    \label{eq:str:beta}
    \beta = \frac{\gamma \alpha_f}{500 G^2 T^{2/3}}= \frac{ \alpha_f}{500 G  T^{2/3} (G+ \alpha_f D)}.
\end{equation}

Finally, we deliver the upper bound of CCV. According to \eqref{eq:cvx:cvv} and $\Phi(x) = x^2 + x$, we have 
\begin{equation}     
    \begin{split}
        \beta^2 Q_T^2 + \tQ_T \leq &  \Reg'_T +  \gamma  \beta GD T \\
        \leq & \frac{224 \beta G^2 }{\gamma  \alpha_f} T^{2 / 3}   (1 + \beta^2 Q_T^2) + \beta G D (\gamma +  2\tQ_T + 1) +  \gamma  \beta GD T \\
        \leq & \frac{1}{2} + \frac{1}{2} \beta^2 Q_T^2 + \tQ_T + \frac{1}{2} (\gamma + 1) +  \gamma  \beta GD  T
    \end{split}     \nonumber
\end{equation}
where the last step is due to \eqref{eq:str:beta} and $2 \beta G D \leq 1$.
Rearranging the above inequality, we have
\begin{equation}
    \begin{split}
        Q_T^2 \leq   \frac{(2 + \gamma)}{\beta^2} + \frac{2 \gamma     GD}{\beta} T = \frac{500 G^2(2 + \gamma)}{\gamma \alpha_f} T^{4/3} + \frac{\gamma^2 \alpha_f D}{250 G} T^{5/3} \quad \Rightarrow  Q_T \leq \O(T^{5/6}).
    \end{split}
\end{equation}

\subsection{Proof of Theorem~\ref{thm:bdt:cvx}}
Similar to the proof of Theorem~\ref{thm:cvx}, we first focus on $\Reg_T'$. Let $\y^{*} = (1- \delta / r) \x^{*}$ and denote $\y_{m(t)}$ as the  auxiliary decision  for $\x_t$.
\begin{equation}
    \begin{split}
        \E \left[\Reg_T'\right] = \underbrace{\E \left[ \sumT \tf_t(\x_t) - \tf_t(\y_{m(t)})\right]}_\ta + \underbrace{\E \left[ \sumT \tf_t(\y_{m(t)}) - \tf_t(\y^{*})\right]}_\tb + \underbrace{\E \left[ \sumT \tf_t(\y^*) - \tf_t(\x^*)\right]}_\tc.
    \end{split}
\end{equation}
Let $t_k$ denote the number of the blocks that uses $G_k$, and $N$ denote the number of blocks that employs different gradient norm estimations. For $\ta$, according to \eqref{eq:bdt:x}, we have 
\begin{equation}    \label{eq:bdt:cvx:ta}
    \begin{split}
        \ta \overset{\eqref{eq:bdt:x}}{=} &\sumT \E\left[\tf_t(\y_{m(t)} + \delta \u_t) - \tf_t(\y_{m(t)})\right] \leq \sum_{m=1}^{T/K} \sum_{j=1}^K \delta  \tG_k \|\u_t\|_2 =  \sum_{k=1}^{N} \delta \tG_k t_k \leq  c \sum_{k=1}^{N}  \tG_k t_k^{3/4}
    \end{split}
\end{equation}
where the first inequality is due to the convexity of $\tf_t(\cdot)$, and  the last inequality is due to $\delta = c T^{-1/4} \leq c t_k^{-1/4}$.

For $\tc$, we have 
\begin{equation}    \label{eq:bdt:cvx:tc}
    \begin{split}
        \tc \leq \sum_{k=1}^{N} \tG_k t_k \|(1- \delta / r) \x^{*} - \x^*\|_2  \leq  \frac{D}{r} \sum_{k=1}^{N}  \delta \tG_k t_k \leq  \frac{cD}{r} \sum_{k=1}^{N}  \tG_k t_k^{3/4}  
    \end{split}
\end{equation}
where the first inequality is due to Assumption~\ref{assump:K-bound}.

Now, we proceed to upper bound $\tb$ and decompose it as below:
\begin{equation}    \label{eq:bdt:cvx:tb}
    \begin{split}
        \tb = \underbrace{\E \left[ \sumT \tf_t(\y_{m(t)}) - \hat{f}_{t, \delta}(\y_{m(t)})\right]}_\td + \underbrace{\E \left[ \sumT \hat{f}_{t, \delta}(\y_{m(t)}) - \hat{f}_{t, \delta}(\y^{*})\right]}_\te + \underbrace{\E \left[ \sumT \hat{f}_{t, \delta}(\y^{*}) - \tf_t(\y^{*})\right]}_\termf,
    \end{split}
\end{equation}
where $\hat{f}_{t, \delta}$ is the smooth version of $\tf_t$ defined in \eqref{eq:smooth_version}.
Then, we introduce the following lemma \cite[Lemma 2.8]{Others:2016:Hazan}
\begin{lemma} 
    \label{lem:2}
  Let $f(\mathbf{x}):\mathbb{R}^{d} \rightarrow \mathbb{R}$ be $\alpha$-strongly convex and $G$-Lipschitz over a convex and compact set $\mathcal{K} \subseteq \mathbb{R}^{d}$. Then, its $\delta$-smooth version defined in \eqref{eq:smooth_version}  has the following properties:
   (i) $\widehat{f}_\delta(\mathbf{x})$ is $\alpha$-strongly convex over $\mathcal{K}_{\delta}$;
   (ii) $| \widehat{f}_\delta(\mathbf{x}) - f(\mathbf{x}) | \leq \delta G$ for any $\mathbf{x} \in \mathcal{K}_{\delta}$;
   (iii) $\widehat{f}_\delta(\mathbf{x})$ is $G$-Lipschitz over $\mathcal{K}_{\delta}$.
\end{lemma}
According to (ii) of Lemma~\ref{lem:2}, we have 
\begin{equation}    \label{eq:bdt:cvx:td}
    \begin{split}
        \td \leq \sum_{k=1}^{N} \delta \tG_k t_k \leq  c \sum_{k=1}^{N}  \tG_k t_k^{3/4}
    \end{split}
\end{equation}
and 
\begin{equation}    \label{eq:bdt:cvx:tf}
    \begin{split}
        \termf \leq \sum_{k=1}^{N} \delta \tG_k t_k \leq  c \sum_{k=1}^{N}  \tG_k t_k^{3/4}.
    \end{split}
\end{equation}

Denote $\hnabla_{t, \delta, m(t)} = \nabla \hf_{t, \delta}(\y_{m(t)})$ and $\y_{m}^* = \argmin_{\y \in \K_\delta}\{F_{b_k:m}(\mathbf{y})\}$. Then, we bound $\te$ in the following way:
\begin{equation}    \label{eq:bdt:cvx:te}
    \begin{split}
        & \te \leq  \E \left[ \sumT \langle \hnabla_{t, \delta, m(t)}, \y_{m(t)} - \y^* \rangle \right] \\
        = & \underbrace{ \E \left[ \sumT \langle \hnabla_{t, \delta, m(t)}, \y_{m(t)} - \y_{m(t)}^* \rangle \right]}_\termg + \underbrace{\E \left[  \sumT \langle \hnabla_{t, \delta, m(t)}, \y_{m(t)}^* - \y_{m(t)+1}^* \rangle \right] }_\termh + \underbrace{ \E \left[ \sumT \langle \hnabla_{t, \delta, m(t)}, \y_{m(t)+1}^* - \y^* \rangle \right] }_{\termi}.
    \end{split}
\end{equation}
Note that $F_{b_k:m}(\mathbf{y})$ is $2$-strongly convex, according to Lemma~\ref{lem:property:str}, we have 
\begin{equation}    \label{eq:termg:1}
    \|\y_{m(t)} - \y_{m(t)}^*\|_2 \leq \sqrt{ F_{b_k:m}(\y_{m(t)}) - F_{b_k:m}(\y_{m(t)}^*)}.
\end{equation}
Therefore, for $\termg$,  we have 
\begin{equation}    \label{eq:bdt:cvx:tg}
    \begin{split}
        \termg \leq & \sum_{m=1}^{T/K} \sum_{j=1}^K  \tG_k \|\y_{m(t)} - \y_{m(t)}^*\|_2 \\
        \overset{\eqref{eq:termg:1}}{\leq} & \sum_{m=1}^{T/K} \sum_{j=1}^K  \tG_k \sqrt{ F_{b_k:m}(\y_{m(t)}) - F_{b_k:m}(\y_{m(t)}^*)} \leq \sum_{m=1}^{T/K} \sum_{j=1}^K  \tG_k \sqrt{\epsilon} \leq \tG_{N} T \sqrt{\epsilon}
    \end{split}
\end{equation}
where the third inequality is due to the stop condition and the last inequality is due to $\tG_k  \leq \tG_{N}$.

To upper bound $\termh$, we introduce the following lemma \cite[Lemma 5]{AISTATS:2020:Garber}.
\begin{lemma}    \label{lem:termh}
    For the block $m$, Algorithm~\ref{alg:bandit:cvx} holds that 
    \begin{equation}    \label{eq:lem:termh}
        \E \left[\|\hnabla_{m}\|_2 \right]^2 \leq  \E \left[\|\hnabla_{m}\|_2^2  \right] \leq K \frac{d^2M^2}{\delta^2} + K^2 \tG_k^2.
    \end{equation}
\end{lemma}

Applying Lemma~\ref{lem:termh}, we have 
\begin{equation}    \label{eq:bdt:cvx:th}
    \begin{split}
        \termh \leq & \sum_{m=1}^{T/K}  \sum_{j=1}^K \tG_k  \E \left[\|\y_{m(t)}^* - \y_{m(t)+1}^*\|_2\right] \\
        = & \sum_{m=1}^{T/K}  \sum_{j=1}^K \tG_k \eta_k \E \left[\|\hnabla_{m}\|_2  \right] \overset{\eqref{eq:lem:termh}}{\leq} \sum_{m=1}^{T/K}  \sum_{j=1}^K \tG_k \eta_k \left(\sqrt{K} \frac{dM}{\delta} + K \tG_k\right)\\
        \overset{\eqref{eq:bdt:eta}}{=} &  \sum_{m=1}^{T/K}  \sum_{j=1}^K \frac{D\sqrt{K}}{ T^{1/2} } + \frac{cD K \tG_k}{ d M  T^{3/4} } \leq  D \sqrt{KT} + \frac{cD K }{ d M  } \sum_{k=1}^N \tG_k t_k^{1/4}
    \end{split}
\end{equation}
where the forth step is due to $\delta = cT^{-1/4}$ and \eqref{eq:bdt:eta}.

Similar to Lemma~\ref{lem:cvx:tb}, we can also obtain that 
\begin{equation}    \label{eq:bdt:cvx:ti}
    \begin{split}
        \termi = &\sum_{k=1}^N \sum_{l=1}^{t_k} \langle \hnabla_{t, \delta, m(t)}, \y_{m(t)+1}^* - \y^* \rangle \leq  \sum_{k=1}^N  \left[ \frac{D^2}{\eta_k} + \eta_k \sum_{m=1}^{t_k} \|  \hnabla_m \|_2^2 \right] \\
        \overset{\eqref{eq:lem:termh},\eqref{eq:bdt:eta}}{\leq} &  \frac{dMD}{c} T^{3/4}\sum_{k=1}^N \tG_k  + \frac{Dd M}{c  } T^{3/4}    + \frac{cD T^{1/2}  }{dM }   \sum_{k=1}^N \tG_k t_k^{1/4},
    \end{split}
\end{equation}
where the inequality is due to $\tG_k = 2^{k-1} \geq 1$ and $K = T^{1/2}$,

By setting  $\epsilon = 4D^2 T^{-1/2}$ and combining \eqref{eq:bdt:cvx:ta}-\eqref{eq:bdt:cvx:te},   \eqref{eq:bdt:cvx:tg}, \eqref{eq:bdt:cvx:th} and \eqref{eq:bdt:cvx:ti}, we obtain 
\begin{equation}
    \begin{split}
        & \E \left[\Reg_T'\right] \\
        \leq & \frac{cD}{r} \sum_{k=1}^{N}  \tG_k t_k^{3/4} +  3c \sum_{k=1}^{N}  \tG_k t_k^{3/4} + \tG_{N} T \sqrt{\epsilon} +  D \sqrt{KT} + \frac{cD K }{ d M  } \sum_{k=1}^N \tG_k t_k^{1/4} \\
         &+  \frac{dMD}{c} T^{3/4}\sum_{k=1}^N \tG_k  + \frac{Dd M}{c  } T^{3/4}    + \frac{cD T^{1/2}  }{dM }   \sum_{k=1}^N \tG_k t_k^{1/4} \\
         \leq & \left( \frac{cD}{r} + 3c  \right)\sum_{k=1}^{N}  \tG_k t_k^{3/4} + 2^{N-1} T^{3/4} + \left(D + \frac{Dd M}{c  } \right) T^{3/4} + \frac{2cD }{dM }  T^{1/2}   \sum_{k=1}^N \tG_k t_k^{1/4} +  \frac{dMD}{c} T^{3/4}\sum_{k=1}^N \tG_k.
    \end{split}
\end{equation}
Note that according to  $\tG_k = 2^{k-1}$ and the Hölder's inequality, we have 
\begin{equation}      
    \begin{split}
         \sum_{k=1}^N \tG_k t_k^{3/4}  \leq   2^{N+2} T^{3/4}, ~
         \sum_{k=1}^N \tG_k t_k^{1/4}  \leq   2^{N+2} T^{1/4}~\text{and}~\sum_{k=1}^N\tG_k \leq 2^{N+2}.
    \end{split}
\end{equation}
Therefore, we obtain 
\begin{equation}
    \begin{split}
        \E \left[\Reg_T'\right] \leq & \left( \frac{cD}{r} + 3c + 1 + \frac{2cD }{dM }  + \frac{dMD}{c} \right)  2^{N+2} T^{3/4}  + \left(D + \frac{d M D}{c  } \right) T^{3/4}  
        \leq   C_1 2^{N+2} T^{3/4} + C_2  T^{3/4}  ,
    \end{split}
\end{equation}
where for brevity, we denote $C_1 = \left( \frac{cD}{r} + 3c + 1 + \frac{2cD }{dM }  + \frac{dMD}{c} \right)$ and $C_2 = \left(D + \frac{d M D}{c  } \right)$.

Recall that in the last block $K$, which employs $\tG_N$, we have 
\begin{equation}
    2^{N-2} = \tG_{N-1} \leq \beta G (\gamma +  \Phi'(\tQ_T)) \leq \tG_N = 2^{N-1},     \nonumber
\end{equation}
which implies that $2^{N+2} \leq 2^4 \beta G (\gamma +  \Phi'(\tQ_T))$. Therefore, we have 
\begin{equation}    \label{eq:bdt:cvx:8}
     \E \left[ \Reg'_T \right] \leq 2^4 \beta G (\gamma +  \Phi'(\tQ_T)) C_1 T^{3/4} + C_2 T^{3/4}.
\end{equation}

Combining \eqref{eq:cvx:regret} and \eqref{eq:bdt:cvx:8}, we obtain 
\begin{equation}
    \E \left[ \Reg_T \right] \leq \frac{ C_3}{\gamma  }  (\gamma +  \Phi'(\tQ_T)) T^{3/4} + \frac{C_2}{\gamma \beta} T^{3/4} - \frac{1}{\gamma \beta} \Phi(\beta Q_T), \nonumber
\end{equation}
where $C_3 = 2^4 G C_1$.
By setting $\Phi(x) = \exp(\lambda x) - 1$ with $\Phi'(x) = \lambda \exp(\lambda x)$, the above inequality can be re-written as
\begin{equation}
    \E \left[ \Reg_T \right]  \leq  C_3 T^{3/4} + \frac{C_2}{\gamma \beta} T^{3/4} + \left[  C_3 \beta T^{3/4}  \lambda -  1\right] \frac{\exp(\lambda \beta Q_T)}{\gamma \beta} \leq  C_3 T^{3/4} +  C_3 C_2  T^{3/4}= \O(T^{3/4}) \nonumber
\end{equation}
where the last step is by setting $\beta  =  C_3^{-1}$, $\gamma = 1$ and $\lambda =   2^{-1} T^{-3/4}$.

Now, we proceed to upper bound CCV. Substituting \eqref{eq:bdt:cvx:8} into \eqref{eq:cvx:cvv}, we have 
\begin{equation}
    \begin{split}
        \Phi(\beta Q_T) \leq  C_3 \beta(\gamma +  \Phi'(\tQ_T)) T^{3/4} + C_2 T^{3/4} + \gamma \beta G D T.  \nonumber
    \end{split}
\end{equation}
By setting $\Phi(x) = \exp(\lambda x) - 1$, $\beta  = C_3^{-1}$, $\gamma = 1$  and $\lambda =  2^{-1}  T^{-3/4}$, the above result delivers
\begin{equation}
    \exp(\lambda \beta Q_T) - 1 \leq (1 + C_2) T^{3/4} + 2^{-1} \exp(\lambda \beta Q^T) + C_3^{-1} GD T,    \nonumber
\end{equation}
which implies 
\begin{equation}
    Q_T \leq 2 C_3 T^{3/4} \ln \left(2 + 2(1 + C_2) T^{3/4} + 2C_3^{-1}GD T\right) \Rightarrow Q_T \leq \O(T^{3/4} \log T). \nonumber
\end{equation}

\subsection{Proof of Theorem~\ref{thm:bdt:str:f}}

First, we focus on upper bounding $\Reg_T$ and introduce the following lemma which is an centralized version of \citet[Theorem 3]{ArXiv:2021:Wan} with $n=1$. 
\begin{lemma}    \label{lem:bdt:str}
    Let  $K=L=T^{2/3}$, and $\delta = c T^{-1/3}$ with $c > 0$ satisfying $cT^{-1/3} < r$, Algorithm~\ref{alg:bandits:str} ensures
    \begin{equation}    
        \E\left[\Reg_T'\right] \leq \frac{1+\log T^{1 / 3}}{\lambda}\left(\frac{8 d^{2} M^{2}}{c^{2}}+8 G'^{2}+3 \lambda^{2} D^{2}\right) T^{2 / 3} + 3 c  G' T^{2 / 3} + \frac{c G' D T^{2 / 3}}{r} + 12 G' D T^{2 / 3}      \nonumber
    \end{equation}    
    where $\lambda = \gamma \beta \alpha_f$ and $G' =  \beta G (\gamma +  \Phi'(\tQ_T))$.
\end{lemma}
According to Lemma~\ref{lem:bdt:str}, we have 
\begin{equation}   
    \begin{split}    \label{eq:bdt:str:1}
        & \E\left[\Reg_T'\right] \\
        \leq &  \frac{1+\log T^{1 / 3}}{\gamma \beta \alpha_f }\left(\frac{8 d^{2} M^{2}}{c^{2}} +3 \gamma^{2} \beta^{2} \alpha_f^{2} D^{2}\right) T^{2 / 3} + \frac{8+8\log T^{1 / 3} + 3c + cD r^{-1}+12D }{\gamma \beta \alpha_f}  G'^{2} T^{2 / 3} \\
        =&  \frac{1+\log T^{1 / 3}}{\gamma \beta \alpha_f }\left(\frac{8 d^{2} M^{2}}{c^{2}} +3 \gamma^{2} \beta^{2} \alpha_f^{2} D^{2}\right) T^{2 / 3} + \frac{ (8\log T^{1 / 3}  + C_3)\beta G^2 }{\gamma \alpha_f} (\gamma + \Phi'(\beta Q_T))^2  T^{2 / 3} \\
        \leq &  \frac{1+\log T^{1 / 3}}{\gamma \beta \alpha_f }\left(\frac{8 d^{2} M^{2}}{c^{2}} +3 \gamma^{2} \beta^{2} \alpha_f^{2} D^{2}\right) T^{2 / 3} + \frac{ (16\log T^{1 / 3}  + 2C_3)\beta G^2 }{\gamma \alpha_f} (\gamma^2 + \Phi'(\beta Q_T)^2)  T^{2 / 3} \\
        = &  \frac{1+\log T^{1 / 3}}{\gamma \beta \alpha_f }\left(\frac{8 d^{2} M^{2}}{c^{2}} +3 \gamma^{2} \beta^{2} \alpha_f^{2} D^{2}\right) T^{2 / 3} + \frac{ (16\log T^{1 / 3}  + 2C_3)\gamma \beta G^2 }{\alpha_f}  T^{2 / 3} +  \frac{ (16\log T^{1 / 3}  + 2C_3)\beta G^2 }{\gamma \alpha_f} \Phi'(\beta Q_T)^2  T^{2 / 3}  
    \end{split}
\end{equation}
where $C_3 = 8 + 3c + cD r^{-1}+12D$ and the second inequality is due to $(a + b)^2 \leq 2(a^2 + b^2)$ for $\forall a, b \in \mathbb{R}$.

Then, we employ the function $\Phi(x) = x^2 $ with $\Phi'(x) = 2x$ and  substitute \eqref{eq:bdt:str:1} into \eqref{eq:cvx:regret}
\begin{equation}
    \begin{split}
        \E\left[\Reg_T\right] = &  (1+\log T^{1 / 3}) \left(\frac{8 d^{2} M^{2}}{c^{2} \gamma^2 \beta^2 \alpha_f   } +3  \alpha_f  D^{2}\right) T^{2 / 3}  + \frac{ (16\log T^{1 / 3}  + 2C_3) G^2 }{\alpha_f}  T^{2 / 3} \\
        & + \left[ \frac{ 8(8\log T^{1 / 3}  + C_3)\beta^3 G^2 }{\gamma  \alpha_f}    T^{2 / 3}  - 1 \right]\frac{1}{\gamma \beta} Q_T^2 \\
        \leq & (1+\log T^{1 / 3}) \left(\frac{8 d^{2} M^{2}}{c^{2} \gamma^2 \beta^2 \alpha_f   } +3  \alpha_f  D^{2}\right) T^{2 / 3}  + \frac{ (16\log T^{1 / 3}  + 2C_3) G^2 }{\alpha_f}  T^{2 / 3}  = \O(T^{2/3} \log T)    \nonumber
    \end{split}
\end{equation}
where the last step is by setting $\beta = 1$ and $\gamma = 16\alpha_f^{-1} (8\log T^{1 / 3}  + C_3) G^2 T^{2 / 3}$.

Next, we consider the CCV and  substitute \eqref{eq:bdt:str:1} into \eqref{eq:cvx:cvv} with the function $\Phi(x) = x^2 $, $\beta = 1$ and $\gamma = 16\alpha_f^{-1} (8\log T^{1 / 3}  + C_3) G^2 T^{2 / 3}$:
\begin{equation}
    Q_T^2 \leq 2C_4  + 2C_5 T^{4/3} + 32\alpha_f^{-1} (8\log T^{1 / 3}  + C_3) G^2 D T^{5 / 3} \Rightarrow Q_T \leq \O(T^{5/6} \log T)    \nonumber
\end{equation}
where 
\begin{equation}
    C_4 = \frac{8d^2M^2(1+\log T^{1 / 3})}{16 (8\log T^{1 / 3}  + C_3) c^2 G^2  }, \quad C_5 = 48 (8\log T^{1 / 3}  + C_3) G^2 D^2 (1 + \log T^{1/3}) + 32 \alpha_f^{-2} (8\log T^{1 / 3}  + C_3)^2 G^4    \nonumber
\end{equation}

\section{Supporting Lemmas}

\subsection{Proof of Lemma~\ref{lem:regret}}
Let $\x^* = \argmin_{\x \in \K} \sumT \tf_t(\x)$, and we have $g_t(\x^*) \leq 0$ for $\forall t \in [T]$. According to \eqref{eq:tf}, it can be verified that 
\begin{equation}    \label{eq:lem:regret:1}
    \tf_t(\x^*) = \gamma \beta f_t(\x^*).
\end{equation}
and $\x^* = \argmin_{\x \in \K} \sumT \tf_t(\x) = \argmin_{\x \in \K} \sumT f_t(\x)$.
Combining \eqref{eq:lem:regret:1} with the definition of $\Reg'_T$, we obtain
\begin{equation}
    \begin{split}
        \Reg'_T = & \sumT \left[ \tf_t(\x_t) - \tf_t(\x^*) \right]  
        \overset{\eqref{eq:tf},\eqref{eq:lem:regret:1}}{=}   \sumT \left[ \gamma \beta f_t(\x) +  \Phi'(\tQ_t) \beta g_t^+(\x) - \gamma \beta f_t(\x^*) \right] \\
         = & \gamma \beta \sumT \left[ f_t(\x) - f_t(\x^*) \right] + \sumT \Phi'(\tQ_t) \beta g_t^+(\x)  
         \overset{\eqref{eq:phi}}{\geq}   \gamma \beta \Reg_T + \sumT \left[ \Phi(\beta Q_t) - \Phi(\beta Q_{t-1}) \right] \\
         = & \gamma \beta  \Reg_T +  \Phi(\beta Q_T),
    \end{split}
\end{equation}
which completes the proof.

\subsection{Proof of Lemma~\ref{lem:cvx:F}}
In this part, we provide a self-contained analysis for Lemma~\ref{lem:cvx:F}, which mainly follows \citet[Lemma 7.4]{Others:2016:Hazan}.
First, we consider the first round in the block $k$, i.e., $t=\ts_k$. Since $\x_{\ts_k} = \x_{\ts_k:\ts_k}^* = \argmin_{\x \in \K} \|\x - \x_{\ts_k}\|_2^2$, we have 
\begin{equation}
   F_{\ts_k:\ts_k-1}(\x_{\ts_k}) -  F_{\ts_k:\ts_k-1}(\x_{\ts_k:\ts_k}^*) = 0 \leq 2D^2 \sigma_{\ts_k, t}. \nonumber
\end{equation}
Then, we assume $F_{\ts_k:t-1}(\x_t) - F_{\ts_k:t-1}(\x_{\ts_k:t}^*) \leq 2D^2 \sigma_{\ts_k, t}$ for any $t \geq \ts_k + 1$, and consider the case with $t+1$:
\begin{equation}    \label{eq:lem:cvx:1}
    \begin{split}
        F_{\ts_k:t}(\x_{t+1}) -  F_{\ts_k:t}(\x_{\ts_k:t+1}^*) = &  F_{\ts_k:t-1}(\x_{t+1}) - F_{\ts_k:t-1}(\x_{\ts_k:t+1}^*) + \eta_k \langle \nabla \tf_t(\x_t), \x_{t+1} - \x_{\ts_k:t+1}^* \rangle \\
        \leq &  F_{\ts_k:t-1}(\x_{t+1}) - F_{\ts_k:t-1}(\x_{\ts_k:t}^*) + \eta_k\langle \nabla \tf_t(\x_t), \x_{t+1} - \x_{\ts_k:t+1}^* \rangle  \\
        \leq & F_{\ts_k:t-1}(\x_{t+1}) - F_{\ts_k:t-1}(\x_{\ts_k:t}^*) + \eta_k \|\nabla \tf_t(\x_t)\|_2 \|\x_{t+1} - \x_{\ts_k:t+1}^* \|_2 \\
        \leq & F_{\ts_k:t-1}(\x_{t+1}) - F_{\ts_k:t-1}(\x_{\ts_k:t}^*) + \eta_k \|\nabla \tf_t(\x_t)\|_2 \sqrt{F_{\ts_k:t}(\x_{t+1}) - F_{\ts_k:t}(\x_{\ts_k:t+1}^*)}
    \end{split}
\end{equation}
The first inequality is by $\x_{\ts_k:t}^* = \argmin_{\x \in \K} F_{\ts_k:t-1}(\x)$, and the last inequality is by the strong convexity of $F_{\ts_k:t}(\cdot)$ with  \eqref{eq:property:str}.

Then, we consider the first term in \eqref{eq:lem:cvx:1}
\begin{equation}    \label{eq:lem:cvx:2}
    \begin{split}
        F_{\ts_k:t-1}(\x_{t+1}) -  F_{\ts_k:t-1}(\x_{\ts_k:t}^*) = & F_{\ts_k:t-1}(\x_{t}+\sigma_{\ts_k,t}\left(\v_{t}-\x_{t}\right)) -  F_{\ts_k:t-1}(\x_{\ts_k:t}^*)  \\
        \leq & F_{\ts_k:t-1}(\x_{t}) -  F_{\ts_k:t-1}(\x_{\ts_k:t}^*) + \langle \nabla F_{\ts_k:t-1}(\x_{t}), \sigma_{\ts_k,t} (\v_{t}-\x_{t})\rangle  +   \sigma_{\ts_k, t}^2 \|\v_{t}-\x_{t}\|_2^2 \\
        \leq &F_{\ts_k:t-1}(\x_{t}) -  F_{\ts_k:t-1}(\x_{\ts_k:t}^*) + \langle \nabla F_{\ts_k:t-1}(\x_{t}), \sigma_{\ts_k, t} (\x_{\ts_k:t}^* - \x_{t})\rangle  +   \sigma_{\ts_k, t}^2 \|\v_{t}-\x_{t}\|_2^2 \\
        \leq & F_{\ts_k:t-1}(\x_{t}) -  F_{\ts_k:t-1}(\x_{\ts_k:t}^*) +  \sigma_{\ts_k, t} (F_{\ts_k:t-1}(\x_{\ts_k:t}^*)- F_{\ts_k:t-1}(\x_{t}))   +   \sigma_{\ts_k, t}^2 \|\v_{t}-\x_{t}\|_2^2 \\
        = &  (1 - \sigma_{ \ts_k, t}) (F_{\ts_k:t-1}(\x_{t}) -  F_{\ts_k:t-1}(\x_{\ts_k:t}^*)) +   \sigma_{\ts_k, t}^2 D^2 \\
        \leq & 2D^2(1 - \sigma_{\ts_k, t})\sigma_{\ts_k, t} + \sigma_{\ts_k, t}^2 D^2 = D^2(2- \sigma_{\ts_k, t}) \sigma_{\ts_k, t}
    \end{split}
\end{equation}
where the first inequality is due to the smoothness of $F_{\ts_k:t}(\x)$, the second inequality is due to \eqref{eq:v} and the third inequality is due to the convexity of $F_{\ts_k:t}(\x)$ and the boundness of $\K$.

Now, we focus on the second term in \eqref{eq:lem:cvx:1}
\begin{equation}    \label{eq:lem:cvx:3}
    \begin{split}
        & \eta_k \|\nabla \tf_t(\x_t)\|_2 \sqrt{F_{\ts_k:t}(\x_{t+1}) - F_{\ts_k:t}(\x_{\ts_k:t+1}^*)} \overset{\eqref{eq:tf:G}}{\leq}  \tG_k \eta_k \sqrt{F_{\ts_k:t}(\x_{t+1}) - F_{\ts_k:t}(\x_{\ts_k:t+1}^*)} \\
        \leq & (\sqrt{D}  \tG_k  \eta_k)^{2 / 3}\left(\frac{ \tG_k   \eta_k}{D}\right)^{1 / 3} \sqrt{F_{\ts_k:t}(\x_{t+1}) - F_{\ts_k:t}(\x_{\ts_k:t+1}^*)} \\
        \leq & \frac{1}{2}(\sqrt{D} \tG_k  \eta_k)^{4 / 3} + \frac{1}{2}\left(\frac{ \tG_k  \eta_k}{D}\right)^{2 / 3}(F_{\ts_k:t}(\x_{t+1}) - F_{\ts_k:t}(\x_{\ts_k:t+1}^*)) \\
        \leq & \frac{1}{8} D^2 \sigma_{\ts_k, t}^2  + \frac{1}{6} \sigma_{\ts_k,t} (F_{\ts_k:t}(\x_{t+1}) - F_{\ts_k:t}(\x_{\ts_k:t+1}^*))
    \end{split}
\end{equation}
where the last step is due to the setting of $\eta_k = \frac{D}{2 \tG_k  T^{3/4} } \leq \frac{D}{2 \tG_k  (t-\ts_k + 1)^{3/4} }$ for all $t \in [T]$ and $\sigma_{\ts_k,t}  = \frac{2}{\sqrt{t - \ts_k + 1}}$.
Substituting \eqref{eq:lem:cvx:2} and \eqref{eq:lem:cvx:3} into \eqref{eq:lem:cvx:1}, we obtain 
\begin{equation}
    \begin{split}
        F_{\ts_k:t}(\x_{t+1}) -  F_{\ts_k:t}(\x_{\ts_k:t+1}^*)  
        \leq &  D^2(2- \sigma_{\ts_k, t}) \sigma_{\ts_k, t}  + \frac{1}{8} D^2 \sigma_{\ts_k, t}^2  + \frac{1}{6} \sigma_{\ts_k:t} (F_{\ts_k:t}(\x_{t+1}) - F_{\ts_k:t}(\x_{\ts_k:t+1}^*)).
    \end{split}
    \nonumber
\end{equation}
Rearranging the above inequality delivers
\begin{equation}    \label{eq:lem:cvx:4}
    F_{\ts_k:t}(\x_{t+1}) -  F_{\ts_k:t}(\x_{\ts_k:t+1}^*) \leq \frac{D^2(2- \frac{7}{8}\sigma_{\ts_k, t}) \sigma_{\ts_k, t} }{1 - \frac{1}{6} \sigma_{\ts_k, t}}. 
\end{equation}
Furthermore, with $\sigma_{\ts_k, t} = 2(t - \ts_k + 1)^{-1/2}$, it is  easy to verify that 
\begin{equation}    \label{eq:lem:cvx:5}
 \frac{ (2- \frac{7}{8}\sigma_{\ts_k, t}) \sigma_{\ts_k, t} }{1 - \frac{1}{6} \sigma_{\ts_k, t}} \leq 2 \sigma_{\ts_k, t+1}.
\end{equation}
We complete the proof by combining \eqref{eq:lem:cvx:4} and \eqref{eq:lem:cvx:5}, as shown below
\begin{equation}
    F_{\ts_k: t}(\x_{t+1}) -  F_{\ts_k: t}(\x_{\ts_k: t+1}^*) \leq 2D^2 \sigma_{\ts_k, t+1}. \nonumber
\end{equation}

\subsection{Proof of Lemma~\ref{lem:cvx:tb}}
% This lemma is summarized from the proof of \citet[Theorem 1]{AAAI:2021:Wan}, and  for completeness, we present the details here. 
First, we  decompose the left side as shown below:
\begin{equation}
   \sum_{j=1}^{t_k} \langle \nabla  \tf_t(\x_t), \x_{\ts_k:t}^* - \x^* \rangle  = \underbrace{\sum_{j=1}^{t_k} \langle \nabla  \tf_t(\x_t), \x_{\ts_k:t}^* - \x_{\ts_k:t+1}^* \rangle}_\tc + \underbrace{\sum_{j=1}^{t_k} \langle \nabla  \tf_t(\x_t), \x_{\ts_k:t+1}^* - \x^* \rangle}_\td,     \nonumber
\end{equation}
where  $\x_{\ts_k:t}^* = \argmin_{\x \in \K} F_{\ts_k:t-1}(\x)$ and  $\x_{\ts_k:t+1}^* = \argmin_{\x \in \K} F_{\ts_k:t}(\x)$. 

Then, we proceed to upper bound $\tc$. Since $F_{\ts_k:t}(\cdot)$ is $2$-strongly convex function, we have 
\begin{equation}
    \begin{split}
        \|\x_{\ts_k:t}^* - \x_{\ts_k:t+1}^*\|_2^2 \leq & F_{\ts_k:t}(\x_{\ts_k:t}^*) - F_{\ts_k:t}(\x_{\ts_k:t+1}^*)\\
        = & F_{\ts_k:t-1}(\x_{\ts_k:t}^*) -  F_{\ts_k:t-1}(\x_{\ts_k:t+1}^*) + \eta \langle \nabla  \tf_t(\x_t), \x_t^* - \x_{t+1}^* \rangle \\
        \leq &  \eta_k \langle \nabla  \tf_t(\x_t), \x_{\ts_k:t}^* - \x_{\ts_k:t+1}^* \rangle \leq \eta \|\nabla  \tf_t(\x_t)\|_2 \|\x_{\ts_k:t}^* - \x_{\ts_k:t+1}^*\|_2 \nonumber
    \end{split}
\end{equation}
where the first step is due to \eqref{eq:property:str} and the second step is due to \eqref{eq:F}.
According to the above inequality, we have 
\begin{equation}
    \|\x_{\ts_k:t}^* - \x_{\ts_k:t+1}^*\|_2 \leq  \eta_k \|\nabla  \tf_t(\x_t)\|_2.    \nonumber
\end{equation}
Therefore, $\tc$ is bounded by 
\begin{equation}    \label{eq:tc}
    \tc \leq \sum_{j=1}^{t_k} \|\nabla  \tf_t(\x_{ t})\|_2 \|\x_{\ts_k:t}^* - \x_{\ts_k:t+1}^*\|_2 \leq  \eta_k \sum_{j=1}^{t_k} \|\nabla  \tf_t(\x_t)\|_2^2.
\end{equation}

Next, to upper bound $\td$, we introduce the following lemma \cite[Lemma 6.6]{Others:2016:Garber}:
\begin{lemma}    \label{lemma:ftrl}
    Let $\{h_t(\x)\}_{t=1}^T$ be a sequence of loss functions and  $\x_t^* \in \argmin_{\x \in \K} \sum_{\tau=1}^t h_\tau(\x)$ for any $t \in [T]$. Then, it holds that
    \[
        \sum_{t=1}^T h_t(\x_t^*) - \min_{\x \in \K} \sum_{t=1}^T h_t(\x) \leq 0.
    \]
\end{lemma}

According to Lemma~\ref{lemma:ftrl}, by setting $h_1(\x) = \eta_k \langle\nabla  \tf_{t}(\x_{t}), \x \rangle +   \|\x - \x_{\ts_k}\|_2^2  $ and $h_t(\x) = \eta_k \langle\nabla  \tf_{t}(\x_{t}), \x \rangle$ for any $t \geq 2$, we have $F_{\ts_k:t}(\x) = \sum_{\tau=\ts_k}^{t} h_\tau(\x)$. Recall that $\x_{\ts_k:t+1}^* = \argmin_{\x \in \K} \{F_{\ts_k:t}(\x) = \sum_{\tau=\ts_k}^{t} h_\tau(\x) \}$. Applying Lemma~\ref{lemma:ftrl}  delivers
\begin{equation}
    \begin{split}
         & \sum_{\tau=\ts_k}^{\ts_k + t_k - 1} h_\tau(\x_{\ts_k:t+1}^*) - \min_{\x \in \K} \sum_{\tau=\ts_k}^{\ts_k + t_k - 1} h_\tau(\x) =   \sum_{j=1}^{ t_k} h_t(\x_{\ts_k:t+1}^*) - \min_{\x \in \K} \sum_{j=1}^{ t_k} h_t(\x)\\ 
        = &  \eta_k \sum_{j=1}^{ t_k} \langle\nabla  \tf_{t}(\x_{t}), \x_{\ts_k:t+1}^* - \hat{\x}^* \rangle + \|\x_{\ts_k: \ts_k+1}^* - \x_{\ts_k}\|_2^2 - \|\hat{\x}^* - \x_{\ts_k}\|_2^2 \leq 0,
    \end{split} \nonumber
\end{equation}
where $\hat{\x}^* = \argmin_{\x \in \K} \sum_{j=1}^{ t_k} h_t(\x)$. Note that $\sum_{j=1}^{ t_k} h_t(\hat{\x}^*) - \sum_{j=1}^{ t_k} h_t(\x^*) \leq 0$.
Therefore, we have 
\begin{equation}    \label{eq:td}
    \begin{split}
        \td = & \sum_{j=1}^{ t_k} \langle\nabla  \tf_{t}(\x_{t}), \x_{\ts_k:t+1}^* - \x^* \rangle = \sum_{j=1}^{ t_k} \langle\nabla  \tf_{t}(\x_{t}), \x_{\ts_k:t+1}^* - \hat{\x}^* \rangle +  \sum_{j=1}^{ t_k} \langle\nabla  \tf_{t}(\x_{t}), \hat{\x}^* - \x^* \rangle \\
        \leq &    \frac{1}{\eta_k}\left( \|\hat{\x}^* - \x_{\ts_k}\|_2^2 - \|\x_{\ts_k: \ts_k+1}^* - \x_{\ts_k}\|_2^2 \right) \leq \frac{D^2}{\eta_k}.
    \end{split}
\end{equation}
Combining \eqref{eq:tc} and \eqref{eq:td} completes the proof.

\end{document}